\documentclass{article}
\usepackage[preprint]{neurips_2025}
\usepackage[utf8]{inputenc} 
\usepackage[T1]{fontenc}    
\usepackage{hyperref}       
\usepackage{url}            
\usepackage{booktabs}       
\usepackage{amsfonts}       
\usepackage{nicefrac}       
\usepackage{microtype}      
\usepackage{xcolor}         

\usepackage{graphicx}
\usepackage{amsmath}
\usepackage{algpseudocode}
\usepackage{algorithm}
\usepackage{multirow}
\usepackage{authblk}
\usepackage{subcaption}

\DeclareMathOperator*{\argmin}{arg\,min}
 
\title{Disentangling Latent Embeddings with Sparse Linear Concept Cones (SLiCC)\thanks{Research was sponsored by the Department of the Navy, Office of Naval Research under ONR award number N00014-24-1-2259. Any opinions, findings, and conclusions or recommendations expressed in this material are those of the author(s) and do not necessarily reflect the views of the Office of Naval Research.}}

\author[1]{Zhi Li}
\author[1]{Hau Phan}
\author[2]{Matthew Emigh}
\author[1,3]{Austin J. Brockmeier}
\affil[1]{Department of Electrical \& Computer Engineering, University of Delaware, USA}
\affil[2]{Naval Surface Warfare Center, Panama City Division, Panama City, Florida, USA}
\affil[3]{Department of Computer \& Information Sciences, University of Delaware, USA}

\begin{document}

\maketitle

\begin{abstract}
Pretrained vision models encode semantic information in latent embedding spaces that are useful for downstream tasks. We hypothesize that this information can be disentangled to separate the content of complex scenes by decomposing each embedding into multiple concept-specific components. We propose Sparse Linear Concept Cones (SLiCC), a supervised dictionary learning approach that estimates a linear synthesis model consisting of sparse, non-negative combinations of groups of dictionary atoms, whose group-wise activity is guided by multi-label information. Each concept-specific component is a non-negative combination of atoms associated with a label and lies in a corresponding concept cone. We introduce an alternating optimization algorithm and an autoencoder formulation of SLiCC. We apply SLiCC to three types of CLIP and DINOv2 embeddings across five datasets. For CLIP, we additionally exploit image-text alignment to enable a self-supervised SLiCC. We also introduce a fully unsupervised autoencoder that enforces group sparsity based on the norm of neuron activations. Quantitative results show that supervised SLiCC improves concept-filtered image retrieval over unfiltered embeddings, including retrieval of finer-grained information not provided during training. Beyond retrieval of individual concepts, the disentangled components enable concept deletion and recombination across images. We further explore conditional image generation using Image-to-Prompt, with qualitative results illustrating retrieval and generation from manipulated concept representations.
\end{abstract}

\section{Introduction}
\label{sec:intro}

Pretrained vision models capture visual and semantic information in dense embedding spaces. These representations are widely applied in tasks such as cross-modal retrieval and conditional generative modeling. In particular, the pre-trained CLIP model (Contrastive Language-Image Pre-Training)~\citep{clip} uses separate image and text encoders to embed inputs into the same vector space using contrastive learning~\citep{gutmann2010noise,poole2019}.  Vision models trained using self-supervised objectives~\citep{chen2020improved,chen2020simple,chen2021empirical,caron2020unsupervised,oquab2023dinov2,van2017neural,esser2021taming,yu2024image,beyer2025highly} also provide semantically rich image embeddings.

Although these embeddings encode information about multiple objects and complex scenes, they are not directly interpretable. For example, consider a query image containing a dog. To retrieve other images containing dogs, we would like to emphasize the ``animal''-related information in the query. Without filtering, an image with a similar overall scene but no dog may rank highly. Thus, \emph{concept-filtered retrieval}, where the user selects which concepts to focus on, has the potential to make image retrieval more flexible and precise. Beyond identifying whether a concept is present, this requires preserving variations within that concept: an ``animal'' representation should retain distinctions that help retrieve similar dogs rather than arbitrary animals.

Towards this goal, we propose Sparse Linear Concept Cones (SLiCC) to disentangle an embedding into concept-associated components, each represented as a non-negative combination of atoms within a concept cone. Given a concept-labeled image dataset, SLiCC learns these cones jointly through supervised dictionary learning. Projecting an embedding onto a selected cone then yields a concept-specific representation for concept-filtered retrieval. We also introduce an autoencoder formulation that directly estimates the coefficients and reconstructs each component by decoding the corresponding latent group. SLiCC does not require text embeddings for supervised training. When image-text alignment is available, as in CLIP, we exploit it to train a self-supervised variant using pseudo-labels obtained through zero-shot classification. Separately, we introduce an unsupervised autoencoder that requires no predefined concept labels and retains the latent groups with the largest coefficient norms for reconstruction.

We evaluate SLiCC across five datasets using three CLIP backbones and one DINOv2 backbone. Supervised SLiCC improves concept-filtered retrieval over unfiltered retrieval at both the general label and finer-grained sub-label levels, despite using only general labels to fit the model. The autoencoder formulation generally provides further improvements, while CLIP-derived pseudo-labels improve retrieval in many settings without image-level training annotations. Our analyses examine the effect of dictionary size on within-concept structure, and ablations assess non-negativity, mean centering, and partial versus full disentanglement. Beyond concept-filtered retrieval, qualitative examples illustrate component subtraction and recombination across images after the disentanglement, with applications of conditional image generation.

\section{Related Work}
 CLIP is leveraged in a number of image retrieval methods using text queries \citep{text_query} or multi-modal  approaches \citep{cir3, cir1, cir2} that use a combination of image and text to retrieve images similar to the contents of a query image modified by the provided query ~\citep{wei2024uniir,jiang2025vlmvec}.  However, directly decomposing image embeddings to perform concept-filtered retrieval remains unexplored.    \citet{attention_head_interpretable} explore the interpretability of the intermediate representations in CLIP, breaking the vision transformer image encoder into patches and attention heads, trying to find the text description for each attention head. However, it does not directly handle decomposition.
 
 SpLiCE~\citep{splice} decomposes an image embedding on a set of fixed word embeddings with sparse coding. It operates in an unsupervised fashion, but the lack of group structure limits the interpretation to instance level.  More similar to our approach, \citet{indirect2022} use PCA to create a subspace with the guidance of a set of word embeddings that describe the same concept or aspect. Compared with these methods, SLiCC operates solely on image embeddings, only using text embeddings to construct pseudo-labels or enable interpretation of the decomposition.

 Underlying SLiCC is a supervised form of constrained dictionary learning, combining non-negativity of coefficients~\citep{ding2008convex} with an assumption of non-overlapping group structure~\citep{szabo2011online,zelnik2012dictionary}.  The group structure makes SLiCC distinct from reconstruction-based disentanglement of latent spaces, such as $\beta$-VAE~\citep{higgins2017betavae} or sparse autoencoders~\citep{sae}, which require post-hoc interpretation. SLiCC is related, but distinct, from other methods for interpreting latent embedding using known (or inferred) concepts. Notably, other work's concepts are often \emph{attributes} of one or more classes, for example `has stripes', whereas we focus on concepts that are often the classes of foreground objects or a classification label of the scene itself (`outdoor' and `indoor').  In particular, SLiCC can be contrasted with Concept Bottleneck Models (CBMs) \citep{koh2020concept} and concept vectors~\citep{kim2018interpretability} by the modeling principle—SLiCC is a generative model of the latent embedding, whereas CBMs and concept vectors are discriminative. CBMs are trained to predict concept occurrence, and their predictions are used for downstream tasks. Concept vectors are simply the normal vector of hyperplanes of linear concept classifiers using the activation of a neural network layer that was directly trained for the downstream task. Interpretable Basis Decomposition \citep{Zhou_2018_ECCV} first learns concept vectors from the penultimate layer, then decomposes the final layer weight for each class  as a non-negative linear combination of these vectors.  Concept whitening~\citep{chen2020concept} is an approach to replace a layer of a neural network to encourage the whitened activation vectors across the neurons to align with a unit vector for the concept, when the concept is active, ensuring the network activations more interpretable. While useful for interpretation,  none of these methods decompose embeddings to capture the variation within a concept. Therefore, they cannot enable concept-filtered retrieval using the original embedding representation of the images in the pool, nor enable conditional generation. 

The methods most similar to SLiCC are Concept Activation Vector (CAV) models that use non-negative matrix factorization (NMF) to learn patterns of the non-negative activation vectors for a given class~\citep{zhang2021invertible,fel2023craft}. SLiCC differs in that it does not require non-negative embeddings, and it explicitly models co-activation of different concepts, enabling disentanglement~\citep{erogullari2025post}.  Our supervised SVD (S-SVD) baseline, which does not model the co-activation of concepts, provides a baseline without disentanglement.

\section{Method}
\label{sec:formatting}

Our proposed method, SLiCC, learns to disentangle the concepts by decomposing embedding vectors into a sparse sum of components, each representative of a concept. We start by formulating the decomposition in the following model.

\subsection{Linear Synthesis Model}

\begin{figure}[htb]
    \centering
    \includegraphics[width=0.4\linewidth, trim=5 50 40 80, clip]{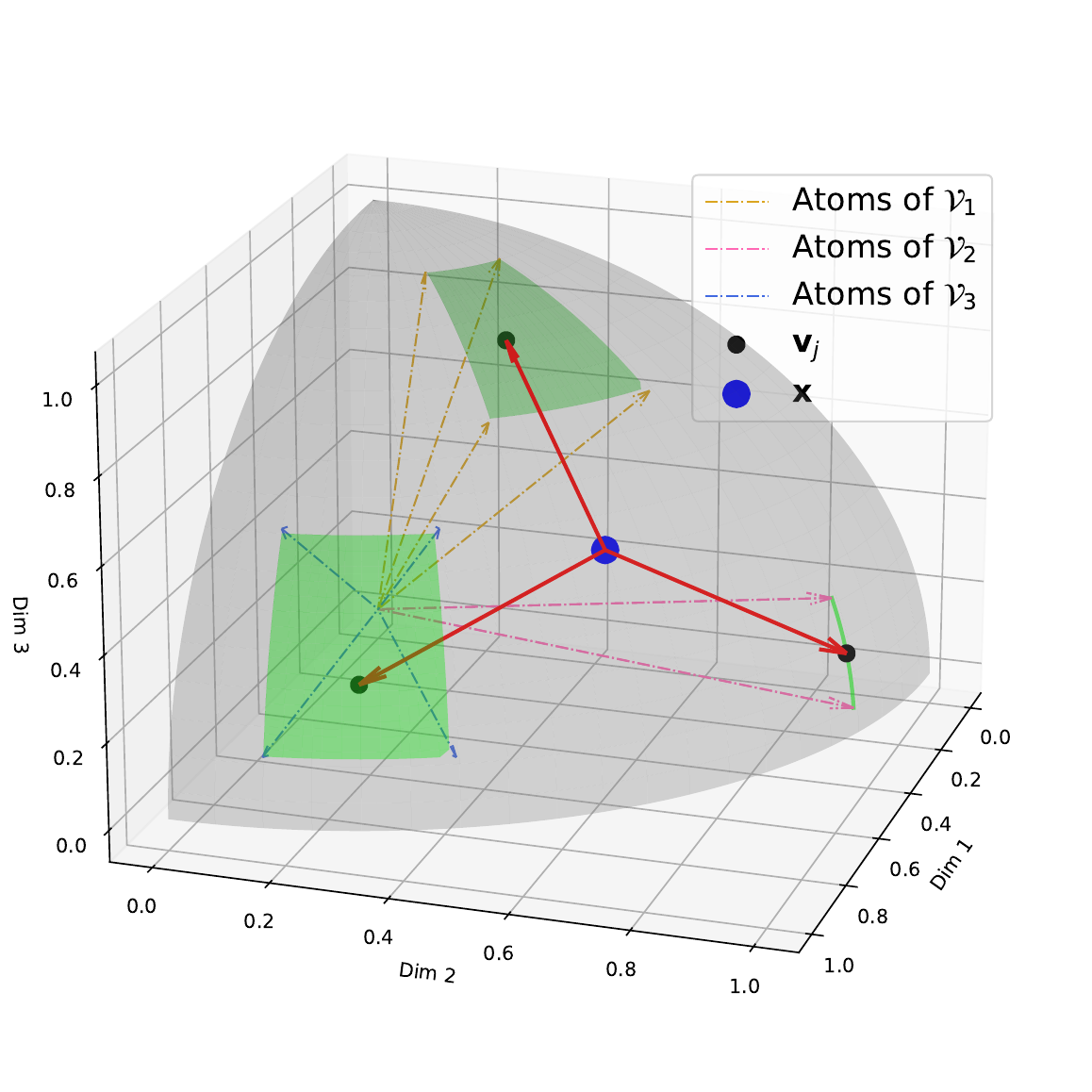}
    \includegraphics[width=0.4\linewidth]{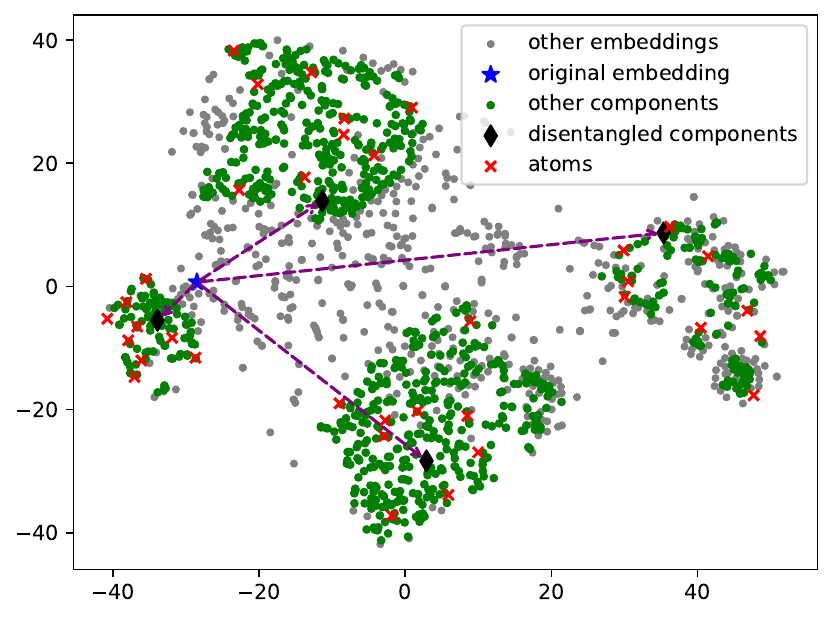}     
    \caption{(Left) Illustration of the disentanglement of a three-dimensional latent space when $S=3$. The gray surface denotes the unit sphere. The concept cones are the green surfaces, defined by 4, 2, and 4 atoms, respectively. The blue dot denotes the original image embedding $\mathbf{x}$. The three black dots denote three components $\mathbf{v}_1, \mathbf{v}_2, \mathbf{v}_3$. (Right) t-SNE visualization of disentangled concept cones by S-SLiCC-AE on ViT-L/14 MS COCO embeddings.}
    \label{fig:method_fig}
\end{figure}

Let $\phi: \mathcal{X} \rightarrow \mathbb{R}^d$ denote a pre-trained neural network that maps an image $X$ to an embedding $\mathbf{x} = \phi(X) \in \mathbb{R}^d$. Let $\mathcal{S}$ be a concept set over $\mathcal{X}$ that determines the semantic information in the image space. If $|\mathcal{S}|=S$, the embedding is approximated as $\mathbf{x} \approx \sum^S_{j=1} \mathbf{v}_j$, where $\mathbf{v}_j \in \mathbb{R}^d$ is the semantic component associated with the concept $j$. By further extending the linear synthesis hypothesis,   each component $\mathbf{v}_j= \mathbf{B}_j\boldsymbol{\alpha}_{j}$ is a linear combination of a set of boundary vectors $\mathbf{B}_j \in \mathbb{R}^{d\times d_j}$, with $d_j$ non-negative coefficients $\boldsymbol{\alpha}_{j}$. Naturally, it is a highly sparse form of decomposition, since $\boldsymbol{\alpha}_{j} \neq \mathbf{0}$---hence naturally, $\mathbf{v}_{j} \neq \mathbf{0}$---if and only if the concept $j$ is represented in the original image. The visualization of the decomposition in Figure\ref{fig:method_fig} shows that $\mathbf{v}_j$ resides in a cone $\mathcal{V}_j$ bounded by $\mathbf{B}_j$.

Here $\boldsymbol{\alpha}_j$ is constrained to be non-negative, which is motivated by the fact that cosine similarity is used by CLIP to measure the similarity of embeddings. Embedding vectors pointing in opposite directions may represent semantic negation or dissimilar input images. Thus, an image embedding $\mathbf{x}$ is considered to semantically include a concept only if its corresponding coefficient is positive.
Note that the embedding of an image with multiple concepts could lie outside of any of the cones its components come from. The t-SNE visualization~\citep{van2008visualizing} of a high-dimensional embedding space in Figure~\ref{fig:method_fig} illustrates this fact.

\subsection{Disentanglement}

Under the linear synthesis hypothesis, the disentanglement of $\mathbf{x}$ requires the learning of $\mathbf{B}$ and $\boldsymbol{\alpha}=[\boldsymbol{\alpha}_j]_{j=1}^S$, and moreover, $\boldsymbol{\alpha}$ is group sparse. Therefore, it is a dictionary learning problem with non-negativity and group sparsity constraint. We consider a training set $\{ X_i\}^N_{i=1}$, and let $\mathbf{X}=[ \mathbf{x}_i ]^N_{i=1} \in\mathbb{R}^{d\times N}$ be the concatenation of the training set mapped to the embedding space. We will now introduce two formulations of an algorithm that performs the disentanglement.

\subsubsection{Sparse Linear Concept Cones (SLiCC)}

For a training set $\{ (X_i,\mathbf{y}_i) \}^N_{i=1}$ containing the concept labels, where $\mathbf{y}_i \in  \{ 0, 1 \}^{S}$ is a $S$-dimensional binary label vector that indicates the presence of each concept in image $X_i$, the problem is
\begin{align}
   \min_{\mathbf{B}, \mathbf{A}\ge 0} \quad & \Vert \mathbf{X} - \mathbf{B} \mathbf{A}\Vert_F^2 \quad
    \textrm{s.t.} \   \boldsymbol{\alpha}_j^i \!= \!\mathbf{0} \quad \textrm{if} \quad \mathbf{y}^i_{j}\! =\! 0. 
\end{align}
where $\mathbf{B}\in\mathbb{R}^{d \times M}$ is the concatenation of $\mathbf{B}_j$, and the activations $\mathbf{A}=[\boldsymbol{\alpha}^i]_{i=1}^N \in \mathbb{R}_{\ge 0} ^{M \times N}$ follow the sparsity pattern indicated by the labels. For simplicity, we assume the number of vectors for each concept is $d_0$ ($M=S \cdot d_0$ total atoms) and vary $d_0$.

We base our supervised SLiCC on the classic dictionary learning algorithm K-SVD~\citep{ksvd}. $\mathbf{B}$ and $\mathbf{A}$ are iteratively updated in K-SVD. The update of $\mathbf{A}$ is straightforward, as we can solve the non-zero part of $\boldsymbol{\alpha}^i$ with non-negative least squares (NNLS). The update of atoms is tricky since in K-SVD it is not guaranteed to obtain non-negative coefficients during this step, leading to it no longer being the optimal solution. In fact, finding an optimal solution with non-negativity—a rank-1 semi-non-negative matrix factorization problem—is NP-Hard~\citep{gillis2015exact}. Nonetheless, we keep the updated atom, and flip it if the negative part of its coefficient has a bigger norm than the positive part, before updating the coefficient again with the new atom (details are given in Section~\ref{sec:optimal_sign}). While this SVD-based update is still a heuristic, the simultaneous update of atom and coefficient converges, as we show in Section~\ref{sec:convergence}. Empirically, the simultaneous update is more efficient. The algorithm is initialized by applying truncated SVD for each class with atoms flipped accordingly. The complete set of steps is described in Algorithm~\ref{alg:supervise_ksvd}.

Although the dictionary learning algorithm is easy to implement and analyze, the time cost becomes expensive when $d_0$ and $S$ grows, especially during the step of updating the atoms. But the algorithm is sufficient to inspire a neural network-based formulation. Let $\mathcal{E}: \mathbb{R}^d \rightarrow \mathbb{R}^{S\times d_0}_{\ge 0}$ be an encoder (linear layer with bias followed by ReLU) that maps the embedding $\mathbf{x}$ to its non-negative coefficient $\boldsymbol{\alpha}$, and $\mathcal{D}: \mathbb{R}^{S\times d_0}_{\ge 0} \rightarrow \mathbb{R}^d$ be a linear decoder that maps the coefficient back to the embedding space.  The training loss is simply $
    1 - \mathrm{sim}(\mathbf{x}, \mathcal{D}(\mathcal{E}(\mathbf{x}) \odot \mathbf{y} ) )
$, where $\mathrm{sim}$ is cosine similarity and $\odot$ is the element-wise multiplication with broadcasting. While the Frobenius norm as reconstruction error is used in our formulation of the dictionary learning, we train the autoencoder to reconstruct the direction of $\mathbf{x}$ since the norm of the reconstructed embedding does not matter in downstream tasks.

\begin{algorithm}[htb]
\caption{Supervised SLiCS dictionary learning}\label{alg:supervise_ksvd}
\begin{algorithmic}
\Require Embedding matrix $\mathbf{X} \in \mathbb{R}^{d \times N}$, cardinality of each dictionary $M_1,\ldots,M_S$ with $M=\sum_{j=1}^S M_j$, concept indices from labels, $\mathcal{I}_j \subset \{1,\ldots, N\}, j\in\{1,\ldots, S\}$, max iterations $T$
\Ensure Dictionary $\mathbf{B}=[\mathbf{B}_j]_{j=1}^S =[\mathbf{b}_m]_{m=1}^{M}$
\For{$j=1$ to $S$}
\State Initialize $\mathbf{B}_j \leftarrow [p_k^* \mathbf{u}^L_k ]_{k=1}^{M_j}$ from the rank-$M_j$ truncated SVD of $[\mathbf{x}_i]_{i\in\mathcal{I}_j} \approx \sum_{k=1}^{M_j} \sigma_k\mathbf{u}^L_k \mathbf{u}_k^{R\top}$ with optimal signs $p_k^*=\begin{cases} 1 & \lVert \max(0,\mathbf{u}^R_k)\rVert_2 \ge  \lVert \max(0,-\mathbf{u}^R_k)\rVert_2 \\ -1 & \text{otherwise}  \end{cases}$ 
\EndFor

\For{$t=1$ to $T$}
\For{$i=1$ to $N$}
\State Solve $\boldsymbol{\alpha}^i = [\boldsymbol{\alpha}^i_j]_{j=1}^S$ using NNLS on supported concepts,  $\boldsymbol{\alpha}_j^i=\mathbf{0}$ if $i\notin\mathcal{I}_j$, as in $\boldsymbol{\alpha}^i = \argmin_{\substack{[\mathbf{a}_j]_{j=1}^S\in \mathbb{R}^M_{\ge0}\\ (1-y_{ij})\lVert \mathbf{a}_j\rVert=0, \  j\in\{1,\ldots,S\} }} \Vert \mathbf{x}_i - \sum_{j=1}^S \mathbf{B}_j \mathbf{a}_j \Vert_2$
\EndFor
\State $\mathbf{A}\leftarrow [\boldsymbol{\alpha}^i]_{i=1}^N$
\For{$m=1$ to $M$}

\State Find the support indices $\mathcal{L}_m \leftarrow  \{l \in \{1,\ldots, M\}: A_{ml} \ne 0\}$ of the $m$th row of $\mathbf{A}$
\State Calculate the residual error without the $m$th atom $\mathbf{E} \gets [\mathbf{x}_l -\sum_{k\neq m} \mathbf{b}_k A_{kl}]_{l\in\mathcal{L}_m}$
\State Obtain rank-1 truncated SVD of residual error  $\mathbf{E}\approx \tilde{\mathbf{b}}_m \tilde{\boldsymbol{\beta}}   $, 
\State Compute the optimal sign $p^\star =  \begin{cases} 1 & \lVert \max(0,\tilde{\boldsymbol{\beta}})\rVert_2 \ge  \lVert \max(0,-\tilde{\boldsymbol{\beta}})\rVert_2 \\ -1 & \text{otherwise}  \end{cases}$
\State Update the atom $\mathbf{b}_m \leftarrow  p^\star\tilde{\mathbf{b}}_m$
\State Update the active coefficients $\mathbf{A}_{m\mathcal{L}_m} \leftarrow   \max(0,p^\star \tilde{\boldsymbol{\beta}}) = \argmin_{\boldsymbol{\beta} \ge \mathbf{0}} \Vert \mathbf{E} - \mathbf{b}_m\boldsymbol{\beta} \Vert_F$
\EndFor
\EndFor
\end{algorithmic}
\end{algorithm}

\subsubsection{Optimal sign for atom update}
\label{sec:optimal_sign}
The optimality of the majority rule follows from the fact that the non-negative least squares problem can be decomposed into solving one entry of $\mathbf{A}_{m\mathcal{L}_m}=[A_{ml}]_{l\in \mathcal{L}_m}$ at a time. Since the solution of non-negative linear regression is given by the active set method~\citep{active_set}, the entry  is either included in the active set if it is non-negative, or not included if it is negative. Hence, the optimal non-negative $\boldsymbol{\beta}^\star$ is obtained by setting the negative coefficient to zero. Additionally, the rank-1 SVD already guarantees the optimal result of ordinary linear regression $\tilde{\boldsymbol{\beta}}$, which allows us to prove that the majority sign rule reduces the error. We begin by decomposing the norm of the error of the rank-1 approximation of the residual,
\begin{equation}
\begin{aligned}\label{eq:ls_expanision}
    \lVert \mathbf{E} - \tilde{\mathbf{b}}_m \tilde{\boldsymbol{\beta}}\rVert_F^2 &= \sum_{l \in \mathcal{L}_m} \lVert \mathbf{e}_l - \tilde{\mathbf{b}}_m \tilde{\beta}_l  \rVert_2^2= \sum_{l \in \mathcal{L}_m} 
    \left ( \lVert \mathbf{e}_l \rVert_2^2   + \underbrace{\lVert \tilde{\mathbf{b}}_m \tilde{\beta}_l  \rVert_2^2}_{\tilde{\beta}_l^2}-2\underbrace{\tilde{\beta}_l\mathbf{e}_l^\top \tilde{\mathbf{b}}_m}_{ \tilde{\beta}_l^2} \right )\\
    &=\sum_{l \in \mathcal{L}_m} \left( \lVert \mathbf{e}_l \rVert_2^2  - \tilde{\beta}_l^2\right) = \sum_{l \in \mathcal{L}_m} \left( \lVert \mathbf{e}_l \rVert_2^2  - \max(0,\tilde{\beta}_l)^2- \max(0,-\tilde{\beta}_l)^2\right)\\
    &=\lVert \mathbf{E}\rVert_F^2 - \lVert \max(0,\tilde{\boldsymbol{\beta}})\rVert_2^2-\lVert \max(0,-\tilde{\boldsymbol{\beta}})\rVert_2^2,
\end{aligned}
\end{equation}
where the simplifications follow from the fact that the optimal least-square coefficient $\tilde{\beta}_l=\mathbf{e}_l^\top \tilde{\mathbf{b}}_m$ is simply the inner-product with the unit vector $\tilde{\mathbf{b}}_m$. This can also be seen from the SVD, since
\begin{equation}
\begin{aligned}
 \left [\tilde{\mathbf{b}}_m^\top \mathbf{e}_l \right]_{l\in \mathcal{L}_m}  &=    \tilde{\mathbf{b}}_m^\top \mathbf{E} 
    = \tilde{\mathbf{b}}_m^\top \mathbf{U}^L\mathbf{\Sigma} \mathbf{U}^{R\top} = \sigma_1 \mathbf{u}_1^{R\top} = \tilde{\boldsymbol{\beta}} \\
    \tilde{\mathbf{b}}_m^\top \mathbf{e}_l &= \tilde{\beta}_l.
\end{aligned}
\end{equation}
For arbitrary sign $p\in \{-1,1\}$, and given the non-negative thresholding operator $(\cdot)_+ = \max(0, \cdot)$, we note that  
\begin{equation}
     \sum_{l \in \mathcal{L}_m} \min_{\beta\ge 0} \lVert \mathbf{e}_l - p\tilde{\mathbf{b}}_m \beta  \rVert_2^2=\sum_{l \in \mathcal{L}_m}  \lVert \mathbf{e}_l - p\tilde{\mathbf{b}}_m (p \underbrace{\mathbf{e}_l ^\top \tilde{\mathbf{b}}_m}_{\tilde{\beta}_l})_+  \rVert_2^2= \lVert \mathbf{E}  -  p \tilde{\mathbf{b}}_m  (p \tilde{\boldsymbol{\beta}})_+ \rVert_2^2,  
\end{equation}
since the optimal non-negative least-square coefficient is the non-negative projection of the optimal least squares component $\beta_l=\max(0,p \mathbf{e}_l^\top \tilde{\mathbf{b}}_m)$. To show the optimality of $p^*$ defined by the majority sign, the non-negative approximation can be expanded as in \eqref{eq:ls_expanision}, yielding 
\begin{equation}
\begin{aligned}
\label{eq:zhi_proof}
    \lVert \mathbf{E} - p\tilde{\mathbf{b}}_m (p\tilde{\boldsymbol{\beta}})_+ \rVert_F^2
    &= \sum_{l \in \mathcal{L}_m} \lVert \mathbf{e}_l - \tilde{p\mathbf{b}}_m (p\tilde{\beta}_l)_+  \rVert_2^2 
    \\&= \sum_{l \in \mathcal{L}_m} 
    \left ( \lVert \mathbf{e}_l \rVert_2^2   + \underbrace{p^2}_{1}\underbrace{\lVert \tilde{\mathbf{b}}_m (p\tilde{\beta}_l)_+  \rVert_2^2}_{(p\tilde{\beta}_l)_+^2\lVert\tilde{\mathbf{b}}_m\rVert_2^2=(p\tilde{\beta}_l)_+^2}-2\underbrace{p(p\tilde{\beta}_l)_+\tilde{\mathbf{b}}_m^\top \mathbf{e}_l}_{p(p\tilde{\beta}_l)_+\tilde{\beta}_l= (p\tilde{\beta}_l)_+^2} \right )\\
    &=\sum_{l \in \mathcal{L}_m} \left( \lVert \mathbf{e}_l \rVert_2^2  - (p\tilde{\beta}_l)_+^2 \right) \\
    &=\lVert \mathbf{E}\rVert_F^2 - \lVert (p\tilde{\boldsymbol{\beta}})_+\rVert_2^2 \\
    &\ge \lVert \mathbf{E}\rVert_F^2 - \max(\lVert (\tilde{\boldsymbol{\beta}})_+ \rVert_2^2, \lVert (-\tilde{\boldsymbol{\beta}})_+ \rVert_2^2) = \lVert \mathbf{E} - p^\star\tilde{\mathbf{b}}_m (p^\star\tilde{\boldsymbol{\beta}})_+ \rVert_F^2,
\end{aligned}
\end{equation}
where the following equality was used, which holds for any scalar $\beta$
\begin{equation}
\begin{aligned}
  p(p\beta)_+\beta= p\beta(p\beta)_+ &= \begin{cases}
            (p\beta)^2, &(p\beta)_+ \ge 0 \\
            0, &(p\beta)_+ < 0
        \end{cases} \\
         &=(p\beta)_+^2.
\end{aligned}
\end{equation}
This shows that $p^\star$ computed by the majority rule is always the optimal sign.

\subsubsection{Convergence}
\label{sec:convergence}  The first step in each iteration, the coefficient update, always reduces the reconstruction error as non-negative linear regression is a convex minimization problem with a unique global minimum.  The update of each atom (and coefficients) can be made contingent on improving the approximation, i.e., the atom update can be skipped if the newly found coefficients and the new atom with optimal sign do not provide a better rank-1 approximation compared to the original atom and previous coefficients. Mathematically, the update improves the approximation if
\begin{equation}
\lVert \mathbf{E} - p^\star\tilde{\mathbf{b}}_m (p^\star\tilde{\boldsymbol{\beta}})_+ \rVert_F^2   =  \sum_{l \in \mathcal{L}_m}  \lVert \mathbf{x}_l -\sum_{k\neq m } \mathbf{b}_k A_{kl}  -   p^*\tilde{\mathbf{b}}_m (p^* \tilde{\beta}_l)_+ \rVert_2^2  \le  \sum_{l \in \mathcal{L}_m}  \lVert \mathbf{x}_l -\sum_{k} \mathbf{b}_k A_{kl} \rVert_2^2.
\end{equation}
With this check, the algorithm is a cyclic minimizer~\citep{cyclic_min}, and thus is guaranteed to converge since both stages reduce the error.

In practice, we let the algorithm proceed with an atom update without checking if it will increase the reconstruction error, as it may end up achieving a better local minimum. 

Finally, we note that an alternative is to use block-coordinate descent on the atom and coefficient update, which will always reduce the error (see Section~\ref{app:alternative}); however, in simulations our simultaneous approach using SVD and thresholding with optimal sign leads to a better approximation.

\subsubsection{Self-supervised and unsupervised SLiCC}

While the labels provide direct guidance on the group sparsity in the supervised cases, we also provide two types of SLiCC when no labels are given. The first one is a self-supervised SLiCC that exploits the image-text alignment of CLIP if the names of the concepts are known. Suppose we have the CLIP text embeddings $\{\textbf{w}_i\}_i^{S}$ of the $S$ concept name words across the templates (for example, ``This is a picture of \{\texttt{word}\}''). The cosine similarity between an image embedding and each text embedding serves as a measurement of how close the word is to the image. Since there are multiple concepts presented in each image we pick the highest $k$ concepts and set them to be active in the pseudo-label. Alternatively, a variable number of active concepts per instance could be selected. Given the pseudo-labels, unsupervised SLiCC uses the same algorithm as the supervised one. We also present an unsupervised variant of SLiCC autoencoder. While the $k$-sparse autoencoder is highly effective for dictionary learning and word semantics disentanglement, its post-hoc probing fails to induce a group-sparse pattern. To address this, similar to our supervised approach, we enforce a group structure on $\mathcal{E}(\mathbf{x}) \in \mathbb{R}_{\ge 0}^{\widetilde{S} \times d_0}$ and only use top-$k$ $\boldsymbol{\alpha}_j$ with the largest norms for the reconstruction. Here $\widetilde{S}$ is the number of unsupervised concepts.

\subsection{Concept-filtered Retrieval}

The disentanglement of $\mathbf{x}$ gives us the semantic components of the original image. Henceforth, we can exploit these components in the embedding space to complete tasks regarding the image on an abstract, semantic level. Instead of the common image retrieval, where the query is considered holistically, only a few of the semantic aspects, or specific concepts, from the query are considered in the concept-filtered retrieval. For the dictionary learning formulation of SLiCC, we perform partial disentanglement to obtain a concept-specific component by approximating $\mathbf{x}$ using only $\mathbf{B}_j$. This projects $\mathbf{x}$ onto the $j$th concept cone, yielding the vector in the cone closest to $\mathbf{x}$, $\tilde{\mathbf{v}}_j = \mathbf{B}_j\tilde{\boldsymbol{\alpha}}_j = \arg\min_{\mathbf{v}\in\mathcal{V}_j}\Vert\mathbf{x}-\mathbf{v}\Vert_2$, where $\tilde{\boldsymbol{\alpha}}_j = \arg\min_{\mathbf{a}\ge\mathbf{0}}\Vert\mathbf{x}-\mathbf{B}_j\mathbf{a}\Vert_2$. Since the concept dictionaries $\{\mathbf{B}_j\}_{j=1}^S$ are generally not mutually orthogonal, the partially disentangled component generally differs from $\mathbf{v}_j$, which accounts for all active concepts. In addition to being simpler, partial disentanglement streamlines the retrieval pipeline, since it requires only the target concept rather than the query's complete label information. Partial disentanglement may also retain information shared across concepts, which, although not semantically exclusive to the target concept, may help preserve similarity to the query. For the autoencoder formulation of SLiCC, we use $\mathbf{v}_j=\mathbf{B}_j\boldsymbol{\alpha}_j=\mathcal{D}(\mathcal{E}(\mathbf{x})\odot \mathbf{m})$ where $\mathbf{m}\in\{0,1\}^M$ has ones at the indices of the $j$th concept.

Accessing latent embeddings to measure the similarity between two images is easy and computationally cheap. Given a query embedding $\mathbf{x}_\star$, concept-filtered retrieval based on the $j$th component uses the similarity score $r_j^i = \mathrm{sim}(\mathbf{v}^{\star}_j, \mathbf{x}_i)$ for ranking a candidate image $\mathbf{x}_i$. The retrieval is based on ranking by sorting the pool of candidate images in descending order based on the similarity scores. As a baseline,  unfiltered  retrieval uses the holistic similarity score $\bar{r}^i =\mathrm{sim}(\mathbf{x}_\star, \mathbf{x}_i)$.

\section{Experiments}

\subsection{Models and Datasets}

We explore three variants of the CLIP model: ResNet-50 ($\mathbf{x} \in \mathbb{R}^{1024}$), ViT-B/32 ($\mathbf{x} \in \mathbb{R}^{512}$) and ViT-L/14 ($\mathbf{x} \in \mathbb{R}^{768}$). The CLIP embeddings are all centered by removing the mean pre-computed the same way as in~\cite{splice} in order to align the distributions of image and text embeddings~\citep{gap}. Note that for zero-shot classification of the self-supervised SLiCC, we use the original embbedings to align with the original paper. We also explore DINOv2 ViT-B/14~\citep{dinov2} ($\mathbf{x} \in \mathbb{R}^{768}$), a self-supervised distilled image embedding model. All embeddings are normalized to lie on the unit sphere.

We use Double MNIST ($S=10$) \cite{double_mnist}, PASCAL VOC 2012 ($S=4$; 20 finer-grained labels in total) \cite{pascal}, MIRFlickr25K~\citep{mirflickr25k} ($S=11$; 6 concepts have finer-grained labels), MS COCO~\citep{coco} ($S=12$; 80 finer-grained labels in total), and a synthetic multi-label CIFAR-100 dataset ($S=20$; 100 finer-grained labels in total) \cite{cifar100}. We use 2,000 training images each for PASCAL VOC 2012 and MIRFlickr25K, and 20,000 each for Double MNIST and MS COCO. For each of these four datasets, we reserve 500 images for validation and 1,000 as queries to evaluate retrieval performance, with the remaining images forming the retrieval pool. For synthetic CIFAR-100, we partition the 60,000 source images into disjoint pools of 10,000, 3,000, 1,500 and 45,500 images to construct the training, query, validation and retrieval sets, respectively. To construct each synthetic image, we choose two or three source images with equal probability, sample them without replacement from the corresponding pool, and place them in randomly selected distinct quadrants of a black canvas. Each synthetic image inherits the union of its constituent coarse and fine labels. This procedure yields 20,000 training images, 1,000 query images, 500 validation images and 40,000 retrieval images. Source images may be reused within a split but are never shared across splits.

\subsection{Retrieval Tasks and Methods}
\label{sec:retrieve}

The goal of the image retrieval is to retrieve images that share the same concepts with the given query. We perform two major types of retrieval task: general label retrieval and sub-label retrieval. Given a query and a presented concept $j$, general label retrieval requires the model to retrieve images that also contain concept $j$. While for sub-label retrieval, the model should retrieve images that contain the same finer-grained concept under $j$. For example, any image with the concept ``animal'' is considered relevant if the query contains the concept ``dog'' in general label retrieval, but an image is only considered relevant in sub-label retrieval if it contains ``dog''. Sub-label retrieval requires higher retrieval precision and tests the capacity of a model to generalize beyond the label information provided during training. The retrieval is quantitatively measured by mean average precision of the top-20 retrieved images (mAP@20). We only quantitatively measure the performance of supervised and self-supervised SLiCC.

\begin{figure*}[t]
    \centering
    \label{figure:tsne_embed}
    \begin{subfigure}[t]{0.32\textwidth}
        \centering
        \includegraphics[width=\textwidth]{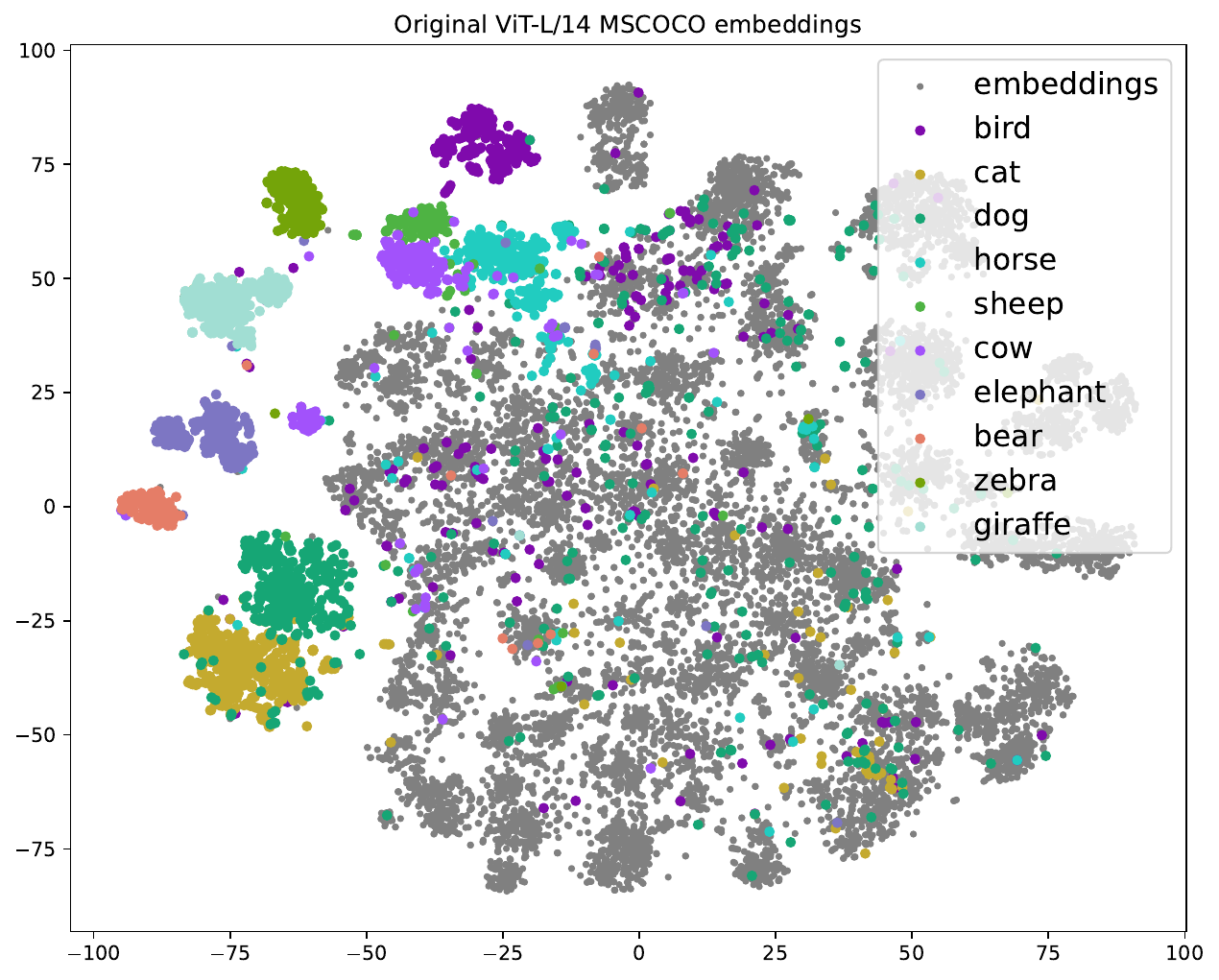}
        \caption{Original embeddings}
        \label{fig:tsne_original}
    \end{subfigure}
    \hfill
    \begin{subfigure}[t]{0.32\textwidth}
        \centering
        \includegraphics[width=\textwidth]{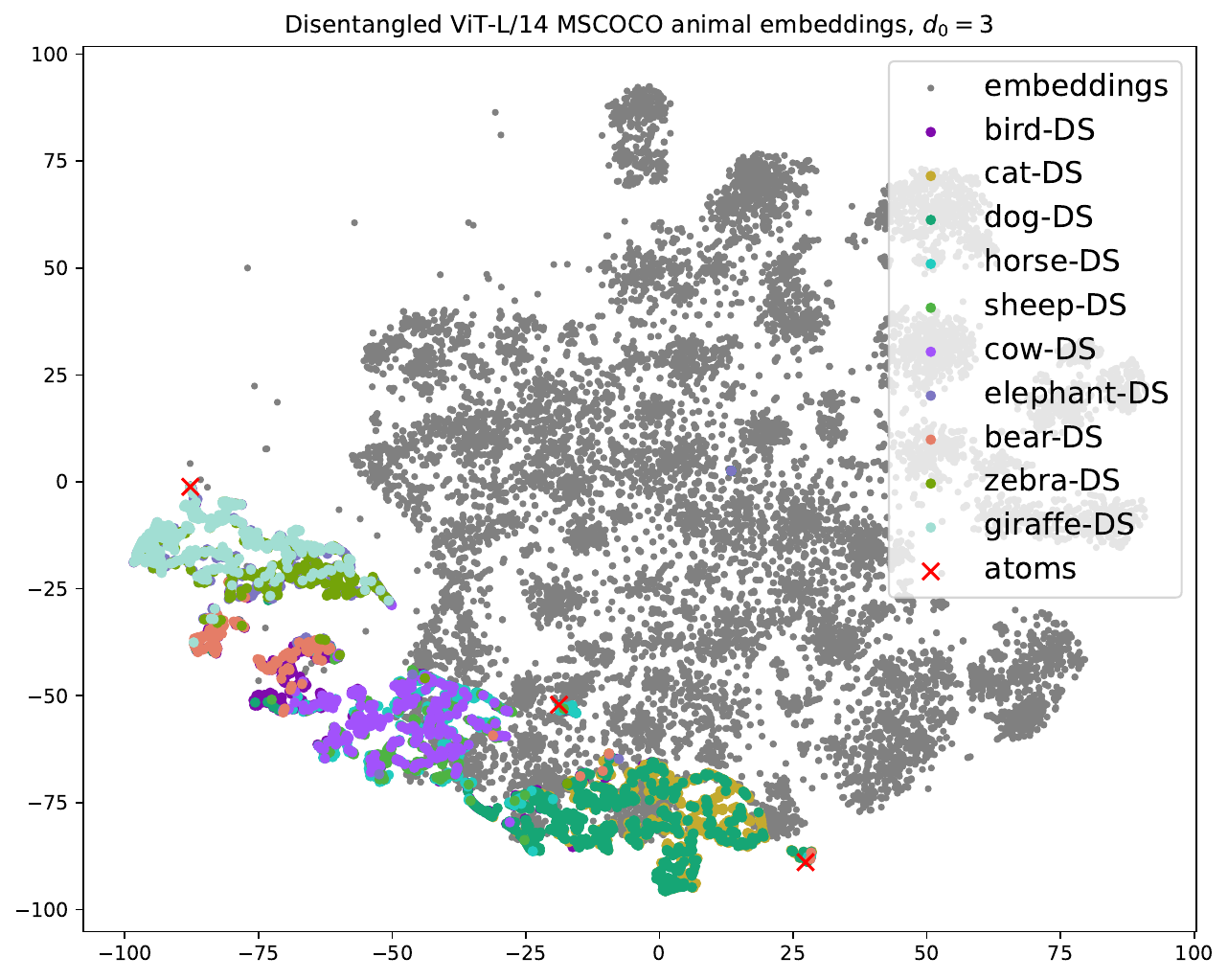}
        \caption{Disentangled, $d_0=3$}
        \label{fig:d0_3}
    \end{subfigure}
    \hfill
    \begin{subfigure}[t]{0.32\textwidth}
        \centering
        \includegraphics[width=\textwidth]{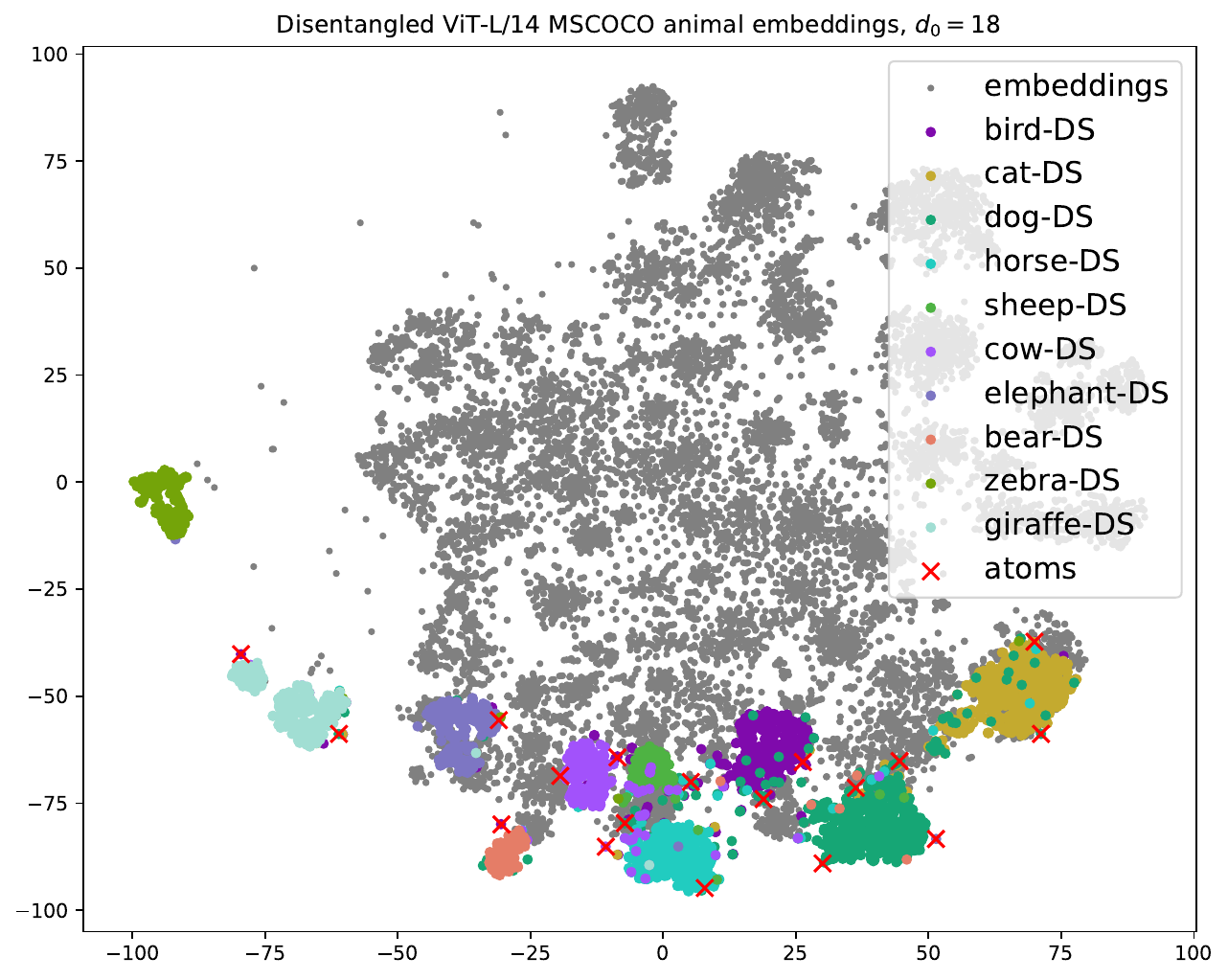}
        \caption{Disentangled, $d_0=18$}
        \label{fig:d0_18}
    \end{subfigure}

    \caption{t-SNE visualizations of embeddings. The gray samples in all three figures are original embeddings with no active ``animal'' concept. The colored samples are either original embeddings with active ``animal'' concept (a), or semantic components disentangled with different $d_0$ (b)(c).}
    \label{fig:tsne_comparison}
\end{figure*}

\paragraph{Hyperparameter Selection}
We first show that the disentanglement pushes embeddings containing similar concepts to lie closer. Figure~\ref{fig:tsne_original} shows that some embeddings with active ``animal'' concept are scattered across the latent space, while with disentanglement, Figure~\ref{fig:d0_18} shows that these embeddings merge with the ``animal'' concept cone. The fact that the cone manifests a cluster structure consistent with the finer-grained labels implies that it preserves the diversity of the concept. Naturally, with a small $d_0$, as shown in Figure~\ref{fig:d0_3}, the cone is limited by the size and forces semantic components to collapse into each other. On the other hand, a large $d_0$ induces redundancy in the cone. Since the ideal cluster structure within the cone is affected by $d_0$, we select $d_0 \in [5, 20]$ in supervised SLiCC by comparing the concept-filtered sub-label retrieval performance on the validation set. We use the same $d_0$ in the corresponding self-supervised SLiCC, and $k$ is estimated by calculating the average number of concepts per sample across the whole dataset. For unsupervised SLiCC, since the $\tilde{S}$ learned groups have no predefined semantic labels, we associate the ground-truth concepts present in each query with distinct latent groups before retrieval. Specifically, we decode and normalize each group separately and compute its cosine similarity with the CLIP text embedding $\{\mathbf{w}_i\}_i^S$. We then perform a greedy exclusive assignment: each unassigned active concept selects its most similar available group; when multiple active concepts select the same group, the concept with the highest similarity is assigned to that group, while the others repeat the selection over the remaining groups. This process continues until all concepts are assigned, ensuring that each group is assigned to at most one concept per query. One group can be associated to different concepts for different queries. The component decoded from the assigned group is then used for concept-filtered retrieval, and we select $k\in\{2,3,4,5\}$, $d_0\in\{8,16,32\}$ and $\widetilde{S}\times d_0\in\{512,1024,2048\}$ by grid search and comparing the concept-filtered sub-label retrieval performance on the validation set.

\paragraph{Settings and Comparisons}
The SLiCC dictionary learning algorithms \textbf{(S-SLiCC)} are trained with 10 iterations. For Double MNIST, MS COCO and synthetic CIFAR100, we run a mini-batch training with a batch size of 2,000. The SLiCC autoencoders \textbf{(SLiCC-AE)} are trained with 150 iterations and a batch size of 64. We compare our methods against two baselines: unfiltered retrieval \textbf{(UF)} and the initialization step of SLiCC dictionary learning algorithms \textbf{(S-SVD)}. We also compare against SpLiCE~\citep{splice}. \textbf{SpLiCE} reconstructs the image embedding from its sparse coefficients for unfiltered retrieval, whereas \textbf{F-SpLiCE} partially reconstructs the embedding using concept-specific groups of word tokens to enable the concept-filtered retrieval. The group structure is constructed by assigning each token to the concept whose text embedding $\mathbf{w}_i$ has the highest cosine similarity with its embedding. We conduct three ablation studies to assess the roles of partial disentanglement, mean centering and non-negativity. First, \textbf{S-SLiCC-FD} reconstructs the concept component $\tilde{\mathbf{v}}_j$ with full disentanglement. Second, \textbf{S-SLiCC-UC} and \textbf{S-SLiCC-AE-UC} use CLIP embeddings without subtracting the mean to evaluate the effect of mean centering on SLiCC. Third, \textbf{S-SLiCC-AE-N} removes the non-negativity constraint on the coefficients, simply using a linear layer as the encoder and allowing signed combinations of dictionary atoms. This tests whether restricting concept-specific components to non-negative combinations improves retrieval.

One key advantage of the concept disentanglement is the ability to manipulate the semantic components of a given embedding. Beyond concept-filtered retrieval, disentangled representations enable direct modification of image embeddings by adding, removing, or recombining specific concepts. Apart from retrieval to demonstrate the effectiveness of the disentanglement, we also demonstrate its application in image generation, as the component $\mathbf{v}_j$ can be used in conditional image generation. Image-to-Prompt~\citep{img2prompt} is a method that leverages Stable Diffusion model conditioned on CLIP ViT-L/14 embeddings, such that the CLIP image embedding is used to generate images similar to the corresponding one to the embedding. Since SLiCC operates on mean-centered embeddings, applying Image-to-Prompt requires additional processing to align the manipulated representation with the uncentered CLIP space expected by the generation pipeline. We add the mean back to the normalized centered embedding $\mathbf{x}_{\text{cen}}$ and renormalize the result to obtain an estimate of the uncentered embedding $\mathbf{x}_{\text{uncen}}$ for Image-to-Prompt. Although the original method operates in the CLIP ViT-L/14 image embedding space, the Procrustes transformation can be used to map an arbitrary latent embedding space into the CLIP ViT-L/14 space.

\subsection{Results}

\begin{figure*}
    \centering
    \includegraphics[height=2.875cm, trim=0 160 0 0,
    clip]{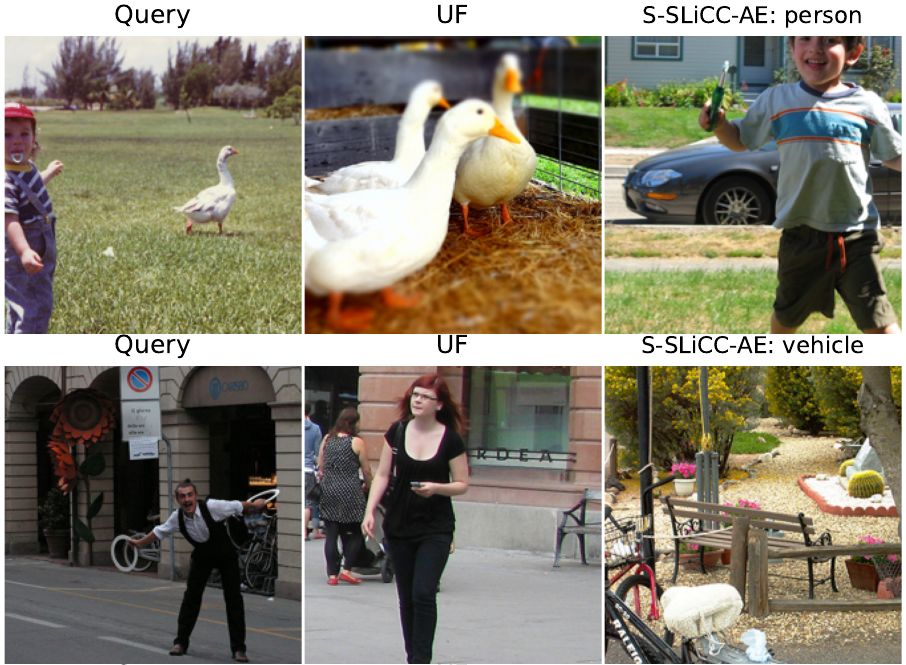}
    \includegraphics[height=2.875cm, trim=0 0 0 160,
    clip]{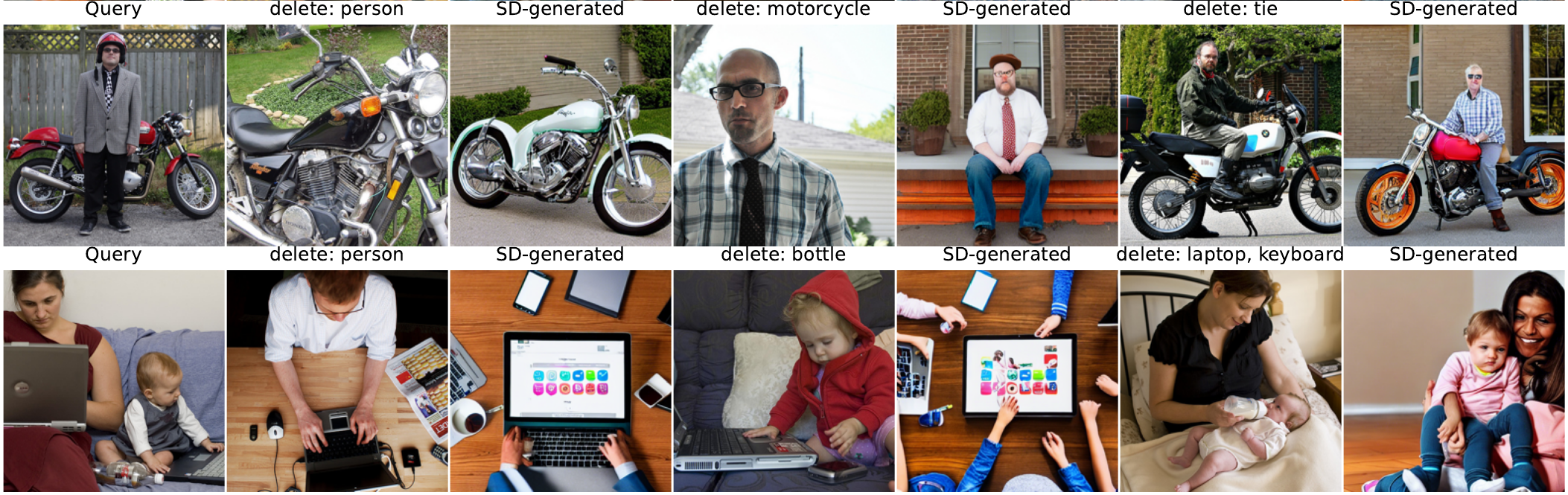}
    \caption{(Left) Top image retrieved by S-SLiCC-AE and UF using DINO embeddings. For the first query, the top two UF results omit the bicycle present in the query's background. For the second query, UF overlooks the child at the edge of the frame. In the two cases, S-SLiCC-AE retrieves relevant images when filtered by the corresponding concepts. (Right) Qualitative results for concept deletion on ViT-L/14 MS COCO embeddings. Concepts are deleted at the general-label level, with the corresponding finer-grained sub-labels indicated in the figure.}
    \label{figure:distractors}
\end{figure*}

\subsubsection{Sanity Check for Concept-filtered Retrieval}
We start by showing qualitative results regarding the original concepts of the query in Figure~\ref{figure:distractors} that proves the necessity of concept-filtered retrieval. As the retrieval results show, unfiltered retrieval ranks the image pool based on the holistic information of the queries, but at the same time, it is possible to ignore some specific parts of the queries that we are interested in. While these concept-specific aspects are encoded in the latent embeddings, the disentanglement is necessary to filter them out, constructing components that preserve the semantic information more clearly. The comparison demonstrates that the selective nature of SLiCC can under certain situations benefit the varying needs of retrieval in terms of different concepts, or semantic information in the query image.

\subsubsection{Quantitative and Qualitative Results}

\begin{figure*}
    \centering
    \includegraphics[height=3.875cm,trim=0 0 300 0,
    clip]{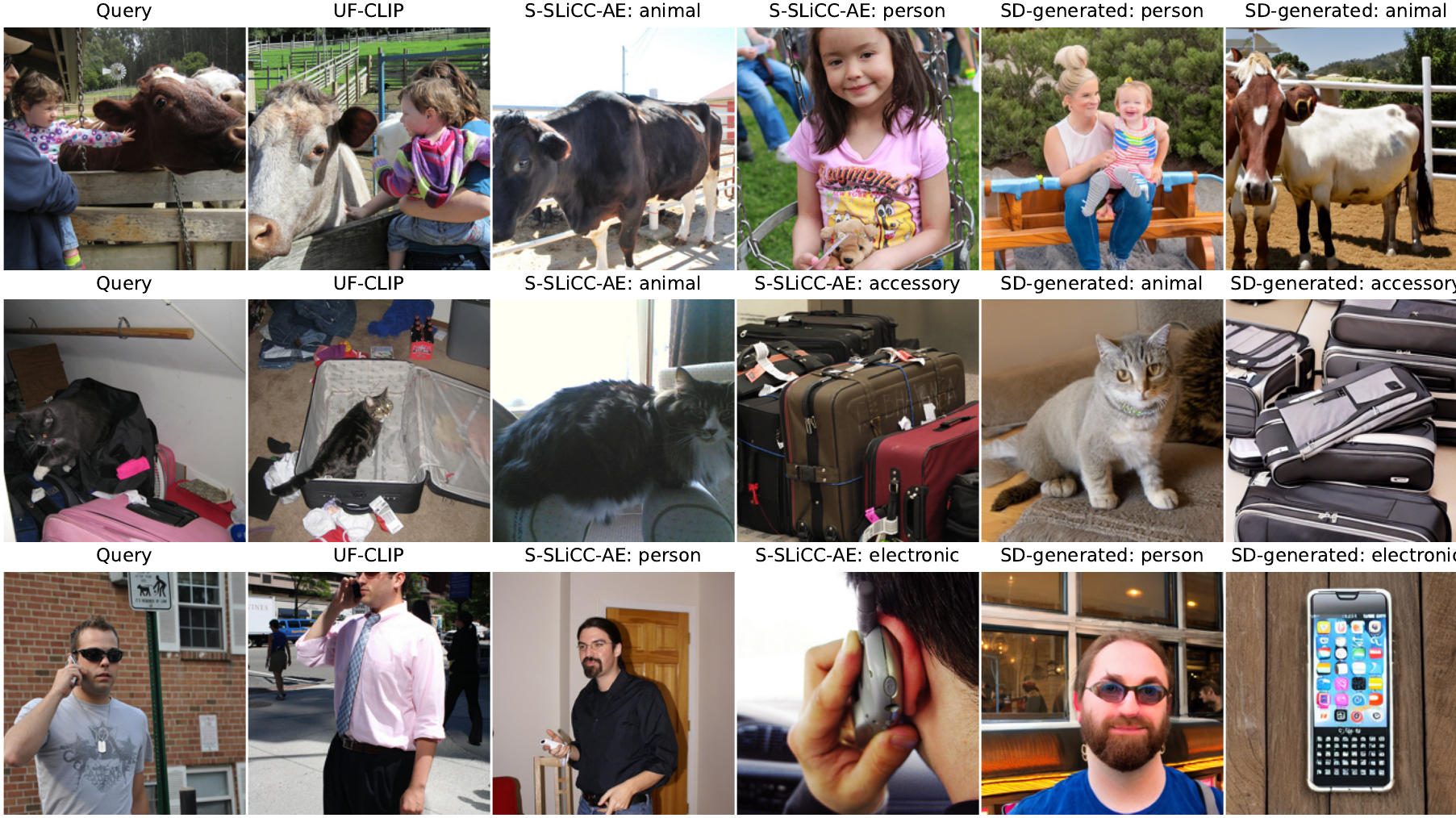}
    \hfill
    \includegraphics[height=3.92cm]{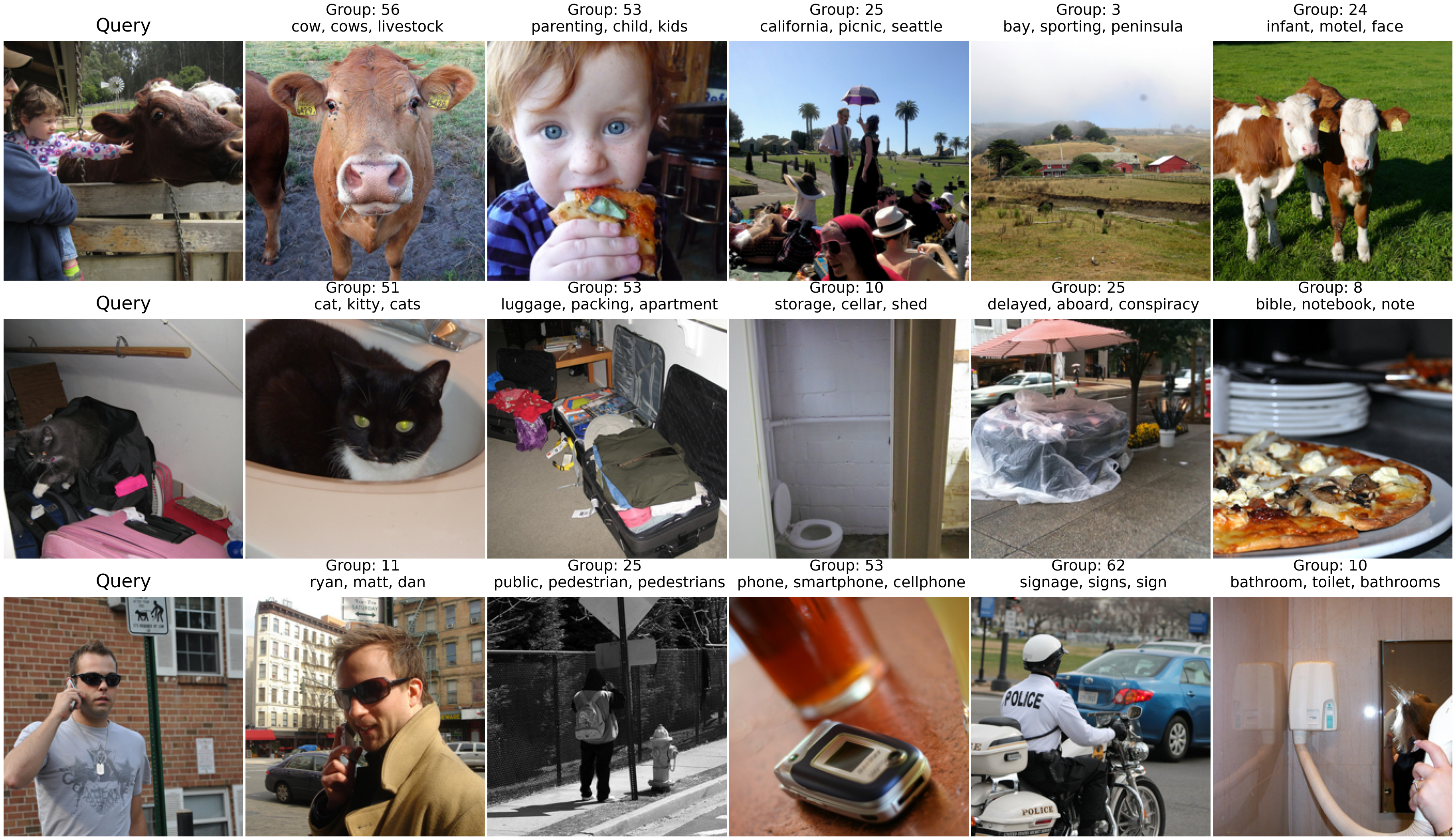}
    \caption{(Left) Qualitative results for S-SLiCC-AE on ViT-L/14 MS COCO embeddings. The first column are queries. The second column are top images retrieved by UF. The third and the fourth columns are top images retrieved by S-SLiCC-AE. The last two columns are Image-to-Prompt generations. (Right) Qualitative results for unsupervised SLiCC on ViT-L/14 MS COCO embeddings. Top-3 word tokens whose embeddings have the highest cosine similarity with the concept component are presented.}
    \label{figure:qual_slicc}
\end{figure*}

Table~\ref{tab:four_tables} reports the quantitative results across five datasets and four type of embeddings. Both S-SLiCC and S-SLiCC-AE consistently outperform the two baselines on general-label and sub-label retrieval, indicating that disentanglement improves the concept-specific retrieval on both the target concepts and their finer-grained categories, despite using only coarse concept labels for training. The autoencoder formulation generally provides further improvements. The ablation results support the constraint of non-negativity and the removal of mean from the image embeddings, although the uncentered embeddings do not consistently perform better on every type of pre-trained model. The results of S-SLiCC-FD show that concept-exclusive information is not sufficient to improve on retrieval, since semantic information not specific to the target concepts can help preserve the similarity with the queries. SS-SLiCC generally performs below its supervised counterpart, although it improves over unfiltered retrieval in many settings, demonstrating the utility of CLIP-derived pseudo-labels when image-level training annotations are unavailable. We complement these quantitative results with qualitative examples of concept-filtered image retrieval using supervised and unsupervised SLiCC-AE in Figure~\ref{figure:qual_slicc}. These examples provide a qualitative assessment of the semantic information obtained and retained by the disentanglement.

\begin{table*}[t]
\centering
\tiny
\setlength{\tabcolsep}{3pt}
\renewcommand{\arraystretch}{1.05}

\begin{minipage}[t]{\textwidth}
\vspace{0pt}
\centering
\resizebox{\linewidth}{!}{%
\begin{tabular}{lccccccccc}
\toprule
\multirow{2}{*}[-0.6ex]{\textbf{CLIP-RN50}}
& \multicolumn{1}{c}{Double MNIST}
& \multicolumn{2}{c}{MIRFlickr25K}
& \multicolumn{2}{c}{MS COCO}
& \multicolumn{2}{c}{Pascal VOC 2012}
& \multicolumn{2}{c}{CIFAR100} \\
\cmidrule(lr){2-2}
\cmidrule(lr){3-4}
\cmidrule(lr){5-6}
\cmidrule(lr){7-8}
\cmidrule(lr){9-10}
& Gen. & Gen. & Sub. & Gen. & Sub. & Gen. & Sub. & Gen. & Sub. \\
\midrule
UF & 0.751 & 0.723 & 0.693 & 0.782 & 0.663 & 0.847 & 0.780 & 0.507 & 0.342 \\
SpLiCE & 0.391 & 0.646 & 0.561 & 0.707 & 0.584 & 0.792 & 0.758 & 0.389 & 0.222 \\
F-SpLiCE & 0.302 & 0.653 & 0.594 & 0.628 & 0.496 & 0.840 & 0.797 & 0.487 & 0.256 \\
S-SVD & 0.948 & 0.909 & 0.633 & 0.908 & 0.674 & 0.922 & 0.300 & 0.717 & 0.252 \\
S-SLiCC & 0.979 & 0.932 & 0.801 & 0.922 & 0.717 & 0.961 & 0.842 & 0.806 & 0.494 \\
S-SLiCC-FD & 0.920 & 0.926 & 0.737 & 0.899 & 0.648 & \underline{0.968} & 0.848 & 0.781 & 0.520 \\
S-SLiCC-UC & \underline{0.980} & \textbf{0.948} & 0.803 & 0.892 & 0.701 & 0.956 & \underline{0.852} & 0.773 & 0.480 \\
S-SLiCC-AE & $\underline{0.980 \pm 0.002}$ & $0.933 \pm 0.004$ & $\underline{0.826 \pm 0.004}$ & $\underline{0.941 \pm 0.002}$ & $\underline{0.746 \pm 0.001}$ & $\mathbf{0.978 \pm 0.001}$ & $\mathbf{0.854 \pm 0.005}$ & $\mathbf{0.847 \pm 0.004}$ & $\mathbf{0.576 \pm 0.004}$ \\
S-SLiCC-AE-UC & $\mathbf{0.985 \pm 0.002}$ & $\underline{0.941 \pm 0.003}$ & $\mathbf{0.836 \pm 0.002}$ & $\mathbf{0.947 \pm 0.003}$ & $\mathbf{0.747 \pm 0.002}$ & $\mathbf{0.978 \pm 0.003}$ & $\underline{0.852 \pm 0.003}$ & $\underline{0.843 \pm 0.002}$ & $\underline{0.575 \pm 0.003}$ \\
S-SLiCC-AE-N & $0.871 \pm 0.031$ & $0.835 \pm 0.030$ & $0.704 \pm 0.026$ & $0.905 \pm 0.015$ & $0.616 \pm 0.014$ & $0.932 \pm 0.016$ & $0.711 \pm 0.026$ & $0.764 \pm 0.020$ & $0.446 \pm 0.006$ \\
SS-SLiCC & 0.709 & 0.859 & 0.720 & 0.882 & 0.651 & 0.936 & 0.797 & 0.707 & 0.341 \\
SS-SLiCC-AE & $0.701 \pm 0.044$ & $0.849 \pm 0.005$ & $0.745 \pm 0.012$ & $0.872 \pm 0.002$ & $0.656 \pm 0.002$ & $0.929 \pm 0.001$ & $0.817 \pm 0.002$ & $0.649 \pm 0.009$ & $0.339 \pm 0.004$ \\
\bottomrule
\end{tabular}%
}
\end{minipage}
\par\vspace{1em}
\begin{minipage}[t]{\textwidth}
\vspace{0pt}
\centering
\resizebox{\linewidth}{!}{%
\begin{tabular}{lccccccccc}
\toprule
\multirow{2}{*}[-0.6ex]{\textbf{CLIP-ViT-B/32}}
& \multicolumn{1}{c}{Double MNIST}
& \multicolumn{2}{c}{MIRFlickr25K}
& \multicolumn{2}{c}{MS COCO}
& \multicolumn{2}{c}{Pascal VOC 2012}
& \multicolumn{2}{c}{CIFAR100} \\
\cmidrule(lr){2-2}
\cmidrule(lr){3-4}
\cmidrule(lr){5-6}
\cmidrule(lr){7-8}
\cmidrule(lr){9-10}
& Gen. & Gen. & Sub. & Gen. & Sub. & Gen. & Sub. & Gen. & Sub. \\
\midrule
UF & 0.688 & 0.728 & 0.692 & 0.791 & 0.671 & 0.848 & 0.770 & 0.533 & 0.362 \\
SpLiCE & 0.356 & 0.669 & 0.592 & 0.737 & 0.627 & 0.800 & 0.769 & 0.442 & 0.280 \\
F-SpLiCE & 0.328 & 0.701 & 0.633 & 0.683 & 0.538 & 0.849 & 0.803 & 0.543 & 0.323 \\
S-SVD & 0.889 & 0.940 & 0.661 & 0.905 & 0.676 & 0.928 & 0.326 & 0.783 & 0.271 \\
S-SLiCC & \underline{0.960} & 0.920 & 0.812 & 0.929 & 0.742 & 0.947 & 0.835 & 0.798 & 0.519 \\
S-SLiCC-FD & 0.868 & 0.911 & 0.766 & 0.910 & 0.663 & 0.967 & 0.844 & 0.790 & 0.562 \\
S-SLiCC-UC & \textbf{0.961} & 0.920 & 0.750 & 0.895 & 0.706 & 0.958 & 0.847 & 0.784 & 0.499 \\
S-SLiCC-AE & $0.945 \pm 0.054$ & $\underline{0.944 \pm 0.003}$ & $\mathbf{0.825 \pm 0.005}$ & $\mathbf{0.959 \pm 0.001}$ & $\mathbf{0.765 \pm 0.001}$ & $\mathbf{0.987 \pm 0.001}$ & $\mathbf{0.851 \pm 0.002}$ & $\mathbf{0.869 \pm 0.003}$ & $\mathbf{0.613 \pm 0.003}$ \\
S-SLiCC-AE-UC & $0.793 \pm 0.111$ & $\mathbf{0.955 \pm 0.003}$ & $\underline{0.824 \pm 0.006}$ & $\underline{0.955 \pm 0.002}$ & $\underline{0.759 \pm 0.007}$ & $\underline{0.981 \pm 0.002}$ & $\underline{0.849 \pm 0.007}$ & $\underline{0.864 \pm 0.001}$ & $\underline{0.606 \pm 0.003}$ \\
S-SLiCC-AE-N & $0.847 \pm 0.047$ & $0.891 \pm 0.020$ & $0.749 \pm 0.015$ & $0.927 \pm 0.013$ & $0.646 \pm 0.011$ & $0.959 \pm 0.014$ & $0.736 \pm 0.023$ & $0.807 \pm 0.012$ & $0.487 \pm 0.015$ \\
SS-SLiCC & 0.236 & 0.887 & 0.741 & 0.887 & 0.673 & 0.930 & 0.805 & 0.696 & 0.382 \\
SS-SLiCC-AE & $0.432 \pm 0.070$ & $0.864 \pm 0.007$ & $0.735 \pm 0.004$ & $0.872 \pm 0.003$ & $0.654 \pm 0.004$ & $0.940 \pm 0.002$ & $0.801 \pm 0.002$ & $0.710 \pm 0.014$ & $0.406 \pm 0.009$ \\
\bottomrule
\end{tabular}%
}
\end{minipage}
\par\vspace{1em}
\begin{minipage}[t]{\textwidth}
\vspace{0pt}
\centering
\resizebox{\linewidth}{!}{%
\begin{tabular}{lccccccccc}
\toprule
\multirow{2}{*}[-0.6ex]{\textbf{CLIP-ViT-L/14}}
& \multicolumn{1}{c}{Double MNIST}
& \multicolumn{2}{c}{MIRFlickr25K}
& \multicolumn{2}{c}{MS COCO}
& \multicolumn{2}{c}{Pascal VOC 2012}
& \multicolumn{2}{c}{CIFAR100} \\
\cmidrule(lr){2-2}
\cmidrule(lr){3-4}
\cmidrule(lr){5-6}
\cmidrule(lr){7-8}
\cmidrule(lr){9-10}
& Gen. & Gen. & Sub. & Gen. & Sub. & Gen. & Sub. & Gen. & Sub. \\
\midrule
UF & 0.883 & 0.714 & 0.680 & 0.794 & 0.694 & 0.851 & 0.784 & 0.601 & 0.450 \\
SpLiCE & 0.417 & 0.665 & 0.591 & 0.747 & 0.659 & 0.794 & 0.783 & 0.498 & 0.358 \\
F-SpLiCE & 0.432 & 0.692 & 0.639 & 0.681 & 0.564 & 0.841 & 0.803 & 0.538 & 0.347 \\
S-SVD & 0.979 & 0.920 & 0.663 & 0.849 & 0.294 & 0.953 & 0.328 & 0.797 & 0.275 \\
S-SLiCC & \underline{0.995} & 0.923 & 0.829 & 0.950 & 0.785 & 0.967 & 0.863 & 0.909 & 0.658 \\
S-SLiCC-FD & 0.979 & 0.906 & 0.806 & 0.938 & 0.739 & 0.977 & \underline{0.880} & 0.911 & 0.721 \\
S-SLiCC-UC & 0.994 & 0.932 & 0.814 & 0.946 & 0.775 & 0.964 & 0.863 & 0.909 & 0.639 \\
S-SLiCC-AE & $\mathbf{0.996 \pm 0.001}$ & $\mathbf{0.958 \pm 0.003}$ & $\mathbf{0.845 \pm 0.007}$ & $\mathbf{0.975 \pm 0.001}$ & $\mathbf{0.810 \pm 0.002}$ & $\mathbf{0.987 \pm 0.001}$ & $\mathbf{0.885 \pm 0.001}$ & $\underline{0.950 \pm 0.002}$ & $\mathbf{0.758 \pm 0.003}$ \\
S-SLiCC-AE-UC & $0.974 \pm 0.042$ & $\underline{0.954 \pm 0.005}$ & $\underline{0.844 \pm 0.006}$ & $\underline{0.972 \pm 0.001}$ & $\underline{0.797 \pm 0.004}$ & $\underline{0.984 \pm 0.001}$ & $0.866 \pm 0.006$ & $\mathbf{0.954 \pm 0.002}$ & $\underline{0.750 \pm 0.003}$ \\
S-SLiCC-AE-N & $0.907 \pm 0.035$ & $0.854 \pm 0.019$ & $0.726 \pm 0.035$ & $0.936 \pm 0.006$ & $0.664 \pm 0.016$ & $0.970 \pm 0.010$ & $0.692 \pm 0.037$ & $0.893 \pm 0.008$ & $0.620 \pm 0.008$ \\
SS-SLiCC & 0.965 & 0.875 & 0.737 & 0.888 & 0.693 & 0.932 & 0.815 & 0.664 & 0.360 \\
SS-SLiCC-AE & $0.882 \pm 0.056$ & $0.892 \pm 0.006$ & $0.734 \pm 0.000$ & $0.891 \pm 0.003$ & $0.689 \pm 0.003$ & $0.927 \pm 0.001$ & $0.810 \pm 0.002$ & $0.748 \pm 0.015$ & $0.460 \pm 0.006$ \\
\bottomrule
\end{tabular}%
}
\end{minipage}
\par\vspace{1em}
\begin{minipage}[t]{\textwidth}
\vspace{0pt}
\centering
\resizebox{\linewidth}{!}{%
\begin{tabular}{lccccccccc}
\toprule
\multirow{2}{*}[-0.6ex]{\textbf{DINO}}
& \multicolumn{1}{c}{Double MNIST}
& \multicolumn{2}{c}{MIRFlickr25K}
& \multicolumn{2}{c}{MS COCO}
& \multicolumn{2}{c}{Pascal VOC 2012}
& \multicolumn{2}{c}{CIFAR100} \\
\cmidrule(lr){2-2}
\cmidrule(lr){3-4}
\cmidrule(lr){5-6}
\cmidrule(lr){7-8}
\cmidrule(lr){9-10}
& Gen. & Gen. & Sub. & Gen. & Sub. & Gen. & Sub. & Gen. & Sub. \\
\midrule
UF & 0.662 & 0.726 & 0.678 & 0.793 & 0.698 & 0.840 & 0.804 & 0.580 & 0.449 \\
S-SVD & 0.837 & \textbf{0.967} & 0.695 & 0.909 & 0.728 & 0.979 & 0.288 & 0.674 & 0.245 \\
S-SLiCC & \underline{0.929} & 0.943 & \textbf{0.818} & \underline{0.984} & \underline{0.785} & 0.972 & \underline{0.858} & 0.918 & 0.675 \\
S-SLiCC-FD & 0.841 & 0.944 & \underline{0.808} & 0.983 & 0.732 & 0.974 & 0.843 & \textbf{0.950} & \textbf{0.761} \\
S-SLiCC-AE & $\mathbf{0.940 \pm 0.004}$ & $\underline{0.966 \pm 0.003}$ & $\underline{0.808 \pm 0.004}$ & $\mathbf{0.990 \pm 0.000}$ & $\mathbf{0.787 \pm 0.003}$ & $\mathbf{0.982 \pm 0.000}$ & $\mathbf{0.866 \pm 0.005}$ & $\underline{0.948 \pm 0.001}$ & $\underline{0.760 \pm 0.002}$ \\
S-SLiCC-AE-N & $0.783 \pm 0.011$ & $0.947 \pm 0.009$ & $0.765 \pm 0.025$ & $0.983 \pm 0.004$ & $0.700 \pm 0.012$ & $\underline{0.980 \pm 0.003}$ & $0.743 \pm 0.045$ & $0.927 \pm 0.004$ & $0.630 \pm 0.012$ \\
\bottomrule
\end{tabular}%
}
\end{minipage}

\caption{Image retrieval quantitative results (mAP@20) for RN50, ViT-B/32, ViT-L/14, and DINO embeddings. \textbf{Gen.}\ and \textbf{Sub.}\ denote general-label and finer-grained sub-label retrieval, respectively. Autoencoder results are reported as mean $\pm$ standard deviation across five runs. Within each type of retrieval for each dataset, the best performance is marked in bold and second best is underlined.}
\label{tab:four_tables}
\end{table*}

\subsubsection{Other Qualitative Results}
\label{sec:other_qual}

In this section we present the qualitative results regarding the manipulated concepts from one or multiple queries. Furthermore, in each part the results are explored from the perspectives of both the image retrieval and image generation. The first type of manipulation is concept deletion. For a query image $X$, we can delete any concept $j$ from its representation and retrieve the images that preserve the semantic information from all but the deleted concept. The representation we use is simply $\mathbf{x}-\mathbf{v}_j$. The second type of manipulation is concept combination. Given a set of query images $\{X_1, X_2, \ldots, X_Q\}$, one or more concepts can be filtered from each image and recombined into a representation containing concept-filtered information across multiple queries. When selecting one concept per query, each concept component $\mathbf{v}^q_{j_q}$ preserves similarity relationships in the embedding space under cosine similarity and can be scaled in the combined representation $\mathbf{z}^\top\mathbf{v}$, where $\mathbf{z}=[z_1,\ldots,z_Q]^\top$ is the scale vector and $\mathbf{v}=[\mathbf{v}^1_{j_1},\ldots,\mathbf{v}^Q_{j_Q}]^\top$ stacks the concept components. Assuming that each component is normalized, equal weights yield a balanced combination, with $\mathbf{z}=0.5\cdot\mathbf{1}$ when $Q=2$.

\begin{figure}[t]
    \centering

\begin{minipage}{\linewidth}
    \centering
    \includegraphics[
        width=0.475\linewidth,
        trim=0 0 541 151,
        clip
    ]{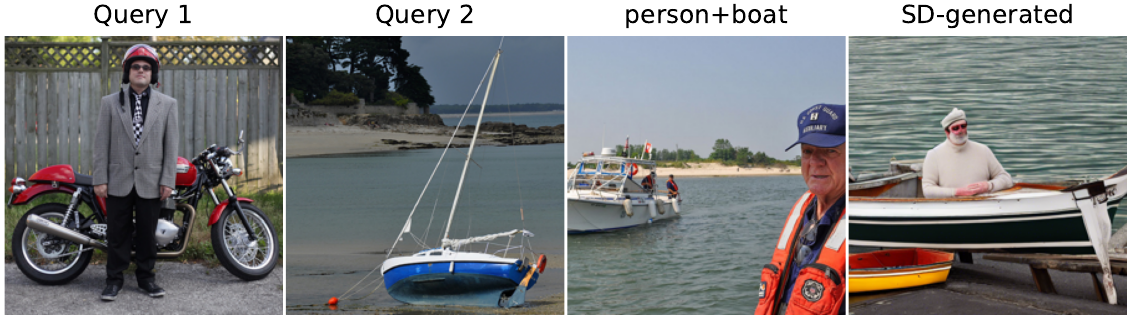}
    \includegraphics[
        width=0.475\linewidth,
        trim=541 0 0 151,
        clip
    ]{figs/combined_concepts_supervised.pdf}
\end{minipage}

\begin{minipage}{\linewidth}
    \centering
    \includegraphics[
        width=0.475\linewidth,
        trim=541 151 0 0,
        clip
    ]{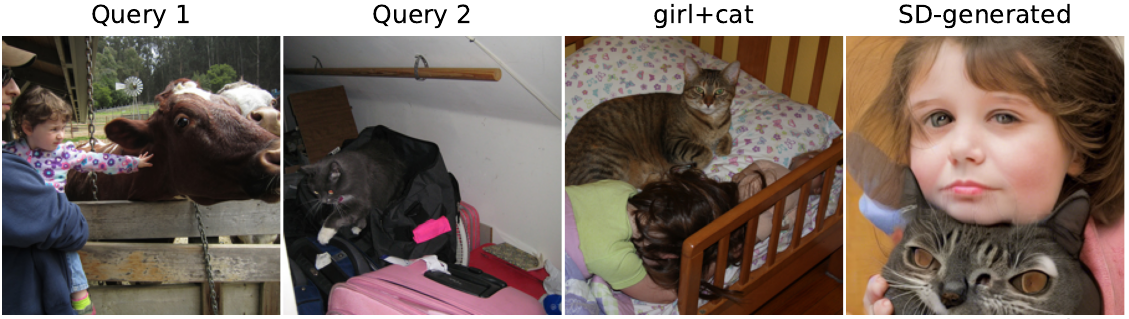}
    \includegraphics[
        width=0.475\linewidth,
        trim=0 0 541 151,
        clip
    ]{figs/combined_concepts_unsupervised.pdf}
\end{minipage}

    \caption{Qualitative results for concept combination on ViT-L/14 MS COCO embeddings. (Top) Disentanglement is performed by S-SLiCC-AE. Concepts are combined at the general-label level, with the corresponding finer-grained sub-labels indicated in the figure. (Bottom) Disentanglement is performed by unsupervised SLiCC-AE. Concepts are identified based on the word tokens presented in Figure~\ref{figure:qual_slicc} (Right).}
    \label{figure:qual_combined}
\end{figure}

However, due to unknown distribution of data in the  high-dimensional embedding space, the image with the highest cosine similarity to an equally weighted combination may not exhibit the desired balance of concepts. We therefore perform retrieval using a set of different scale vectors $\mathbf{z}$. For each top-retrieved image $\mathbf{x}'_{\mathbf{z}}$, we estimate a representative scale vector by solving $\mathbf{z}^\star=\arg\max_{\tilde{\mathbf{z}}}\mathrm{sim}(\mathbf{x}'_{\mathbf{z}},\tilde{\mathbf{z}}^\top\mathbf{v})$, holding the retrieved image fixed. For $Q=2$, this simplifies to $z^\star=\arg\max_{\tilde{z}\in[0,1]}\mathrm{sim}(\mathbf{x}'_{\mathbf{z}},\tilde{z}\mathbf{v}^1_{j_1}+(1-\tilde{z})\mathbf{v}^2_{j_2})$, where $\mathbf{z}^\star=[z^\star,1-z^\star]^\top$. We then rank the top-retrieved images in ascending order of $|z^\star-0.5|$ to favor balanced combinations. The representative scale vector associated with the top-ranked image is used for image-to-prompt visualization. If multiple representative scale vectors are associated with the same top-ranked image, we select the one whose corresponding original retrieval scale $z$ yields the highest similarity $\mathrm{sim}(\mathbf{x}'_{\mathbf{z}},z\mathbf{v}^1_{j_1}+(1-z)\mathbf{v}^2_{j_2})$, where $z$ belongs to the set of scales used for retrieval. The retrieval results and the generation results are shown in Figure~\ref{figure:distractors} and Figure~\ref{figure:qual_combined}.

\section{Conclusion}
SLiCC enables the decomposition of image embeddings into meaningful concept subspaces. The methods uses dictionary learning and formulation of autoencoders with either supervision or self-supervision from zero-shot classification. The immediate benefit of the decomposition is increased effectiveness of concept-filtered image retrieval and concept-specific conditional generation. For CLIP, SLiCC also enables interpretable descriptions in both supervised and unsupervised cases.
We found that DINOv2~\citep{dinov2} that uses self-supervised training, and CLIP ViT-L/14, captured rich information from images and offered the best concept-filtered retrieval performance.
The qualitative results including the combination and deletion of concepts further demonstrate the disentanglement. The unsupervised SLiCC explores the ability of the model to capture the semantic information from a dataset without guidance.
Our concept-level disentanglement indicates that non-negative, group-sparse linear synthesis model is a promising and viable perspective to understand the latent embeddings of vision-language models and other deep neural networks.

{
\small
\bibliographystyle{plainnat}
\bibliography{li_bib}
}


\appendix
\section{Alternative update of atoms}
\label{app:alternative}
In K-SVD, using SVD to simultaneously update coefficients and atoms during the atom update step is optimal to reduce the norm of the error in the residual. However, with an additional non-negative constraint, it is not optimal, as the problem is a rank-1 semi-NMF problem~\citep{gillis2015exact}. Rather than our heuristic, which uses the thresholded coefficients after identifying the optimal sign, an alternative is to perform block-coordinate descent by updating atom and coefficients separately. Given the previous coefficients, each entry of the atom can be optimized in parallel by ordinary least squares. Then, given updated atom, the non-negative linear regression can be used to obtain the updated coefficients, again in parallel:
\begin{equation}
        \acute{\mathbf{b}}_m = \argmin_{\boldsymbol{\gamma}} \lVert  \mathbf{E} - \boldsymbol{\gamma} \boldsymbol{\beta}_0   \rVert_F^2, \quad \quad
        \acute{\boldsymbol{\beta}} = (\breve{\boldsymbol{\beta}})_+ ,\quad \breve{\boldsymbol{\beta}}= \argmin_{\boldsymbol{\beta}} \lVert  \mathbf{E} - \acute{\mathbf{b}}_m\boldsymbol{\beta}  \rVert^2_F,
\end{equation}
where $\boldsymbol{\beta}_0= [A_{ml}]_{l\in \mathcal{L}_m}$ from the previous coefficient update. Both of these problems have closed-form solutions, $\acute{\mathbf{b}}_m=\mathbf{E} \boldsymbol{\beta}_0^\top (\boldsymbol{\beta}_0 \boldsymbol{\beta}_0^\top)^{-1}=\dfrac{\mathbf{E} \boldsymbol{\beta}_0^\top}{\lVert   \boldsymbol{\beta}_0  \rVert^2}$ and $\breve{\boldsymbol{\beta}} = (\acute{\mathbf{b}}_m^\top \acute{\mathbf{b}}_m)^{-1} \acute{\mathbf{b}}_m^\top \mathbf{E}=\dfrac{\acute{\mathbf{b}}_m^\top \mathbf{E}}{\lVert\acute{\mathbf{b}}_m \rVert^2}$, respectively. With the inclusion of a normalizing step to enforce $\acute{\mathbf{b}}_m$ to have unit norm, the solutions are scaled yielding
\begin{equation}\label{eq:both}
        \acute{\mathbf{b}}_m =  \dfrac{\mathbf{E} \boldsymbol{\beta}_0^\top}{\lVert   \mathbf{E} \boldsymbol{\beta}_0^\top  \rVert}, \quad         \quad \breve{\boldsymbol{\beta}} = (\acute{\mathbf{b}}_m^\top \acute{\mathbf{b}}_m)^{-1} \acute{\mathbf{b}}_m^\top \mathbf{E} = \dfrac{\boldsymbol{\beta}_0 \mathbf{E}^\top \mathbf{E}}{\lVert \boldsymbol{\beta}_0 \mathbf{E}^\top  \rVert}.
\end{equation}

Following Eq.~\ref{eq:zhi_proof}, the residual error after the update is
\begin{equation}
\label{eq:error_alt}
        \lVert \mathbf{E} - \acute{\mathbf{b}}_m (\breve{\boldsymbol{\beta}})_+ \rVert_F^2 = \lVert \mathbf{E}\rVert_F^2 - \lVert (\breve{\boldsymbol{\beta}})_+\rVert_2^2.
        \end{equation}
The residual error can be lower bounded by upper bounding the second term as 
\begin{equation}\label{eq:zhi_proof2}
    \begin{aligned}
    \rVert (\breve{\boldsymbol{\beta}})_+ \rVert_2^2 
    &\le \rVert \breve{\boldsymbol{\beta}} \rVert_2^2 = \dfrac{\lVert \boldsymbol{\beta}_0 \mathbf{E}^\top \mathbf{E} \rVert^2_2}{\lVert \boldsymbol{\beta}_0 \mathbf{E}^\top  \rVert^2_2} 
    &\le \left \lVert  \dfrac{\boldsymbol{\beta}_0 \mathbf{E}^\top }{\boldsymbol{\beta}_0 \mathbf{E}^\top} \right \rVert_2^2 \cdot  \lVert  \mathbf{E}  \rVert_\mathrm{op}^2 = \sigma_1^2 = \lVert \tilde{\boldsymbol{\beta}} \rVert^2_2.
    \end{aligned}
\end{equation}

Although it is hard to compare between $\lVert (\breve{\boldsymbol{\beta}})_+ \rVert$ and $\lVert (p^*\tilde{\boldsymbol{\beta}})_+ \rVert$, we know from \eqref{eq:zhi_proof2} that the total norm of $\breve{\boldsymbol{\beta}}$ is smaller than $\tilde{\boldsymbol{\beta}}$, making it less likely to reduce more residual error. Furthermore, it also shows that at the beginning phase of the iteration, where $\acute{\mathbf{b}}_m^\top$ is far apart from the space described by $\mathbf{E}$, the last inequality will be stronger. In fact, it is easy to show that when the true optimum is close, i.e., when $\mathbf{E}$ can be well-approximated by a rank-1 matrix, these two methods are equal. Assuming $\mathbf{E} = \sigma_1 \mathbf{u}_1^L \mathbf{u}_1^{R\top}$, since $\acute{\mathbf{b}}_m^\top$ needs to lie in the space of $\mathbf{E}$, which only has one left singular vector, leading to $\acute{\mathbf{b}}_m^\top = \mathbf{u}_1^{L\top}$, then
\begin{equation}
\begin{aligned}
    \breve{\boldsymbol{\beta}} &= \sigma_1 \mathbf{u}_1^{L\top} \mathbf{u}_1^L \ \mathbf{u}_1^{R\top} = \sigma_1 \mathbf{u}_1^{R\top} = \tilde{\boldsymbol{\beta}}.
\end{aligned}
\end{equation}
Considering that new coefficients can be written in terms of previous coefficients, given the initial update of non-negative set of coefficients, $\acute{\boldsymbol{\beta}}_{1} = (\breve{\boldsymbol{\beta}})_+$, a heuristic approach is to iteratively update,
\begin{equation}\label{eq:alt_update}
        \acute{\boldsymbol{\beta}}_{t+1} =  \left ( \dfrac{\acute{\boldsymbol{\beta}}_{t} \mathbf{E}^\top \mathbf{E}}{\lVert \acute{\boldsymbol{\beta}}_{t} \mathbf{E}^\top  \rVert}\right)_+,
\end{equation}
for $t\in\{1,\ldots,T\}$ as a modified power method.

\section{Additional Results}\label{app:additional_results}
\subsection{Interpretation via Word Captions}
A set of words are selected to interpret the cones learned by SLiCC. We consider the single English word token from the captions of LAION-400m dataset, and follow~\citet{splice} to filter out any NSFW words and pick the most frequent 10,000 words. The text embeddings are normalized, mean-centered, and then re-normalized. Top 5 words that have the smallest non-negative linear reconstruction error using the concept dictionary $\mathbf{B}_j$ are selected to describe the cone $\mathcal{V}_j$ since they are the closest to the cone in the embedding space.

 The word captions of S-SLiCC-AE and SS-SLiCC-AE with ViT-L/14 MS COCO embeddings are presented in Table~\ref{tab:coco_captions}. As shown in the tables, both supervised and self-supervised cone disentanglement shows semantic consistency. Although for SS-SLiCC-AE, there is less inconsistency for some concepts. For example, the ``electronic'' concept is described by ``clock'' and ``train''.
 
\begin{table}[tbh]
\centering
\caption{Cone captions of MS COCO on CLIP ViT-L/14 embeddings. ``Concept Cone'' column shows the ground truth concept label. Words are ordered by the ascending reconstruction error.}
\label{tab:coco_captions}
{\scriptsize
\begin{tabular}{lll}
\hline
Concept Cone   & S-SLiCC-AE word caption                          & SS-SLiCC-AE word caption                             \\ \hline
person     & granddaughter, girl, participants, woman, grandson & granddaughter, girl, grandma, woman, herself \\
vehicle    & bike, bikes, boat, boats, motorcycle              & truck, buses, automotive, vehicle, motorcycle \\
outdoor    & benches, bench, nyc, brooklyn, lights             & street, beach, lake, nature, nyc \\
animal     & dogs, horses, horse, pups, cows                   & bear, dogs, pups, bears, horse \\
accessory  & rainy, rain, rainfall, umbrellas, rains           & luggage, umbrellas, suitcase, umbrella, rainy \\
sports     & skate, baseball, tennis, skateboard, surf         & baseball, tennis, surfers, skate, surf \\
kitchen    & drink, beverage, drinks, wine, drinking           & kitchen, cooking, kitchens, pantry, chef \\
food       & pizza, donuts, cake, donut, veggies               & pizza, donuts, donut, cake, sandwiches \\
furniture  & bed, lounge, room, toilet, beds                   & bathroom, bed, toilet, workspace, furniture \\
electronic & cellphone, phone, smartphone, smartphones, phones & clock, clocks, trains, train, plane \\
appliance  & fridge, kitchen, bathroom, oven, microwave        & fridge, plumbing, appliance, refrigerator, firefighter \\
indoor     & clock, books, library, bookshelf, toothbrush      & winery, indoors, wines, indoor, wine \\ \hline
\end{tabular}}
\end{table}

\subsection{Atom Co-occurrence versus Atom Similarity}
\begin{figure}[htb]
    \centering
\includegraphics[width=0.39\linewidth,clip,trim={27cm 9cm 0 0}]{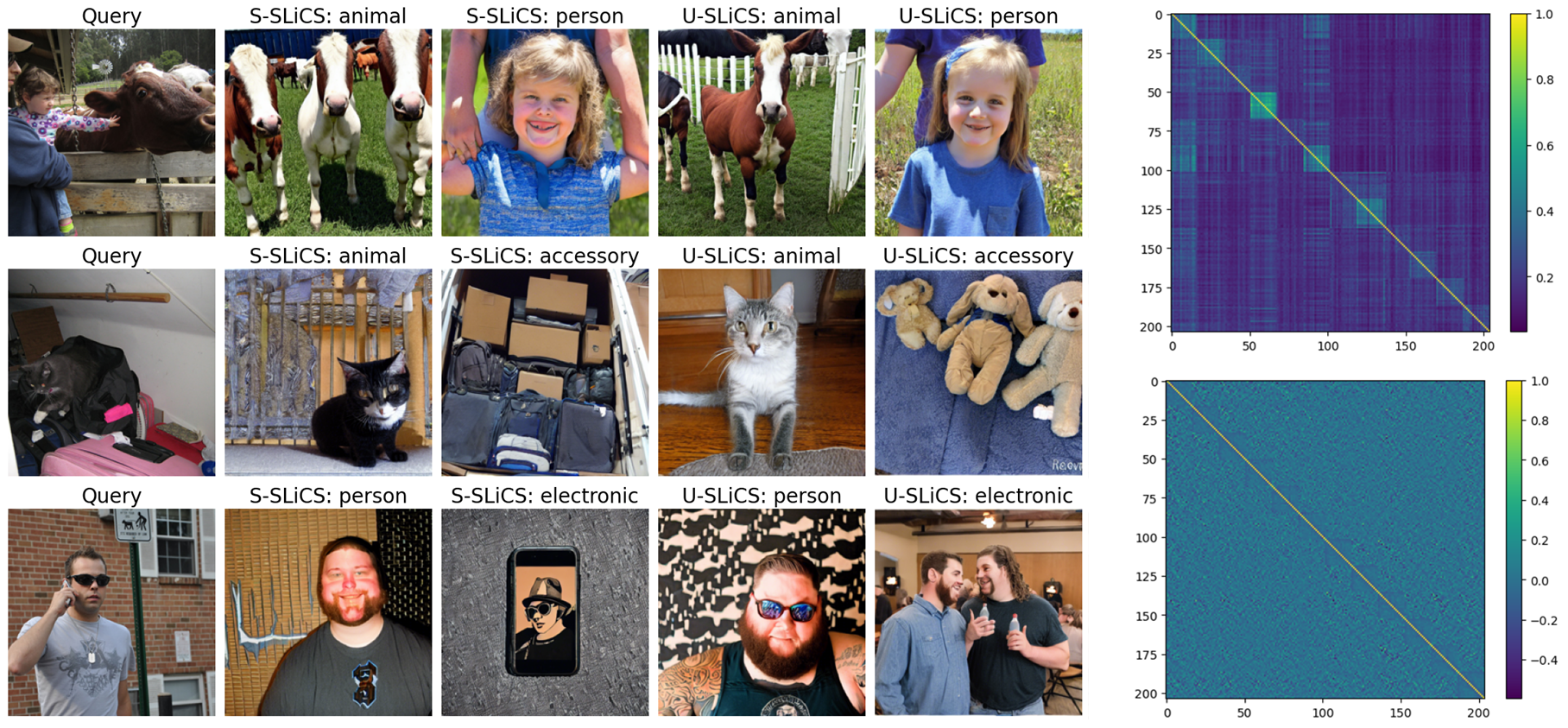}
    \includegraphics[width=0.39\linewidth,clip,trim={27cm 0 0 9cm}]{figs/vit_qual_visual.png}
    \caption{Relationship between the atoms across the concept dictionaries  and their coefficients in supervised SLiCC. Dimensionality chosen by retrieval performance on validation set (see Section~\ref{sec:retrieve}).  (Left) the co-occurrence matrix of active atoms for images in the validation set. Each row is normalized separately so that the diagonal entries are 1. Within each group the co-occurrence rate is generally  higher than between groups. (Right) The cosine similarity matrix of the atoms in the same dictionary. The within-group similarity ($-0.056 \pm 0.062$) is slightly lower than cross-group similarity ($0.000 \pm 0.134$).}
    \label{fig:matrices}
\end{figure}

To emphasize that the group structure creates atoms that work together, coefficients are estimated without using the concept labels using  greedy sparse coding (Orthogonal Matching Pursuit~\citep{omp1, omp2} with non-negative constraint) across the validation set examples. Then the co-occurrence between atoms is calculated from the frequency that one atom co-occurs with another based on both having non-zero coefficients.
Figure~\ref{fig:matrices} shows that the group structure is evident from the co-occurrence matrix, whereas the atoms within a concept group are nearly orthogonal (as seen by the cosine similarity). We also confirmed that there is no redundancy of atoms for the same concept in that none are interior to the convex hull for all the tested values of $d_0$.
During training, atoms associated to one concept are always allowed to co-occur compared to atoms from other concepts; thus, they tend to be nearly orthogonal to better capture the diversity within a concept.

\subsection{Diffusion Posterior Sampling}
\label{sec:dps}

Diffusion posterior sampling (DPS)~\cite{chung2022diffusion} is a method to sample from the posterior distribution $p_{X|Y=y}$ using the stochastic denoising (also known as reverse) process of a diffusion model as a generative prior $p_{X}$ and a measurement likelihood function $\mathcal{L}(x|y)=p_{X=x|Y}(y)$, where $X$ is a generated image and $y=\mathcal{A}(x_0)$ is a known forward measurement. Assuming isotropic Gaussian noise on the measurement,  the log likelihood is $\log \mathcal{L}(x|y)=\mathit{Constant}-\dfrac{1}{\sigma^2} \lVert y-\mathcal{A}(x)\rVert_2^2$. The discrete-time sampling process iterates from $t=T$ to 0, where the variable $x_t$ denotes the noised image and $\hat{x}_0(x_t)=\mathbb{E}[x_0|x_t]$ denotes the expected denoised version under the model. The DPS algorithm combines a denoise step $x'_{t-1}\leftarrow f_{t}(x_t,z_t)$, where $f_t$ incorporates the denoising model with the scheduled noising with $z_t$, with a gradient ascents step of the log-likelihood function 
\begin{equation}
    x_{t-1} \leftarrow x_t' -\zeta \nabla_{x_t}\lVert y-\mathcal{A}(\hat{x}_0(x_t))\rVert_2,
\end{equation}
where $\zeta$ is a fixed step size that absorbs the forward measurement noise as well as normalizing the gradient of the squared norm by the value $||y-\mathcal{A}(\hat{x}_0(x_t))||_2$ as implemented by \citep{chung2022diffusion}. We note that this corresponds to a Laplacian noise model $\log \mathcal{L}(x|y)=\mathit{Constant}-\zeta  \lVert y-\mathcal{A}(x)\rVert_2$.

As part of the SLiCC framework, we propose a novel forward measurement model to sample from a concept-specific cone. In this case, $Y\in\{1,\ldots,S\}$ denotes the concept of interest ($y=j$), and $\log \mathcal{L}(x|j)=\mathit{Constant}-\zeta \lVert \phi(x)-\mathcal{G}_j(\phi(x))\rVert_2$,
where $\phi$ is the pre-trained embedding network (with $\ell_2$-normalization) and $\mathcal{G}_j$ is an implicit function that project's $\phi(x)$ to $\tilde{v}_j=\argmin_{\mathbf{v}\in\mathcal{V}_j} \Vert \mathbf{x}_\star -  \mathbf{v} \Vert_2 $. Clearly, the log-likelihood function is maximized if the embedding lies with the  positive cone such that it can be fully explained by $\mathcal{B}_j$ and decreases with increasing distance. By using this measurement model, we can sample images from the concept-specific cones to understand the SLiCC model.
 \begin{equation}
    x_{t-1} \leftarrow x_t' - \zeta  \nabla_{x_t}\lVert \phi(\hat{x}_0(x_t))-\mathcal{G}_j(\phi(\hat{x}_0(x_t)))\rVert_2,
 \end{equation}

Using the DINOv2 ViT-B/14 embeddings and the SLiCC model trained on MS COCO, we generated 5 images for each fo the 12 concepts, with $\zeta=2$. The posterior samples are presented in Figure~\ref{fig:dps-2}. We include the codes of visualization based on the original implementation in the supplementary material.
\begin{figure}[htb]
    \centering
    \includegraphics[width=1\linewidth]{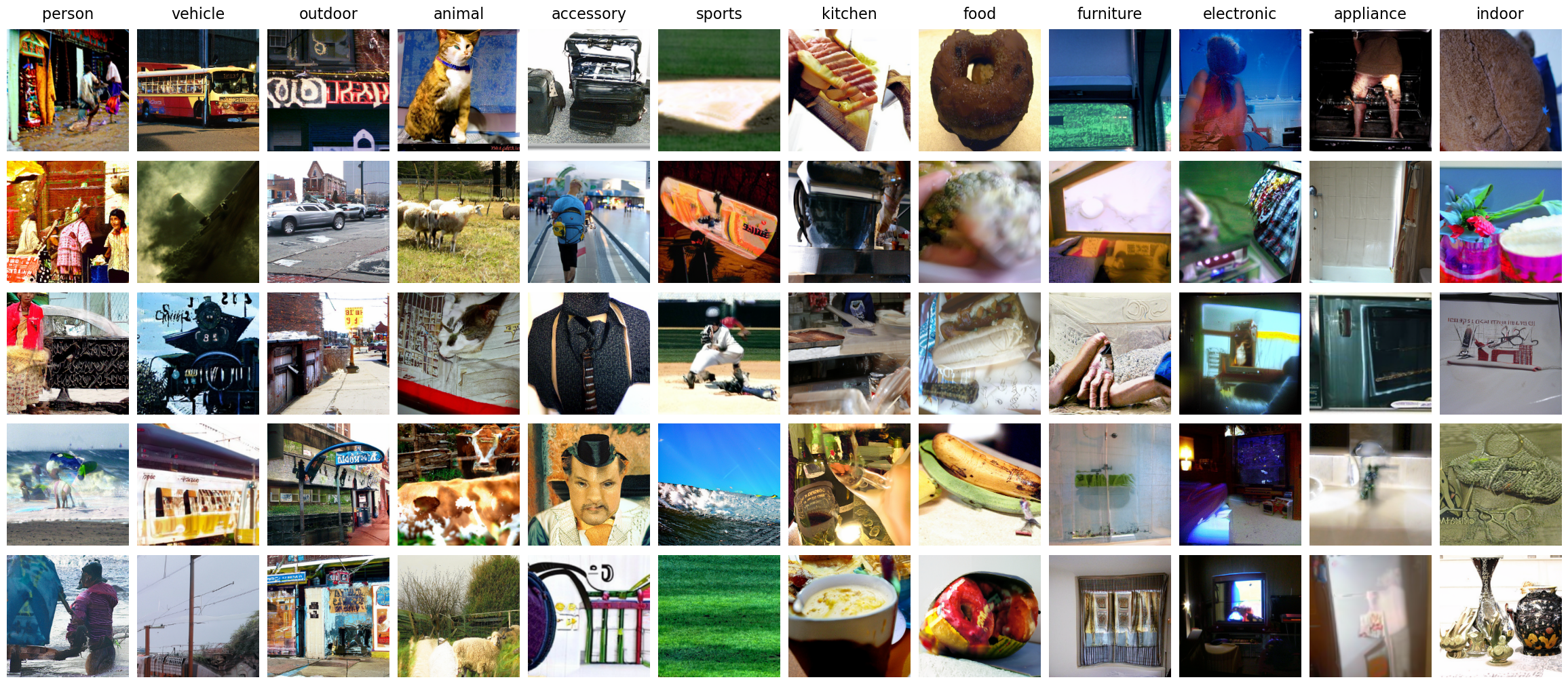}
    \caption{The DPS sampling results from the cone with DINOv2 ViT-B/14 embeddings (MS COCO). Each columns contains five samples of one concept.}
    \label{fig:dps-2}
\end{figure}

Although this approach encourages posterior samples to come from the cone, it does not consider whether the coefficients are prototypical. A more nuanced approach would be to model the conditional distribution of coefficients given the concept using naive Bayes with a simple exponential distribution as done in prior work~\cite{grosse2012shift}. The resulting likelihood would encourage the generation of images with coefficients for the atoms to match the model.

\section{Comparison of Concept Combination Strategies}
\label{app:concept_combination}

We compare the concept combination strategy described in \ref{sec:other_qual} with two baselines to assess the balanced combination of concepts. Given two target concepts, we independently sample with replacement from the validation set one query image containing each concept and repeat this procedure to obtain 100 query pairs. All methods use the same query pairs and image pool. The target finer-grained sub-labels are taken from the source queries.

\paragraph{Unfiltered embedding combination}
The first baseline directly combines the original query embeddings without disentanglement. Given a query pair $\mathbf{x}_1$ and $\mathbf{x}_2$, the combined query is
$
(\mathbf{x}_1+\mathbf{x}_2)/
\|\mathbf{x}_1+\mathbf{x}_2\|_2.
$
This baseline retains the full scene information from both source images, including content unrelated to the selected concepts, and tests whether combining the original representations is sufficient for the retrieval task.

\paragraph{Average concept combination}
The second baseline uses SLiCC to disentangle the concept components from the queries. Given a query pair $\mathbf{x}_1$ and $\mathbf{x}_2$, the concept components are $\mathbf{v}_{j_1}^1$ and $\mathbf{v}_{j_2}^2$. Then the combined query is $
(\hat{\mathbf{v}}_1+\hat{\mathbf{v}}_2)/
\|\hat{\mathbf{v}}_1+\hat{\mathbf{v}}_2\|_2,
$
where $\hat{\mathbf{v}}_j=\mathbf{v}_j/\|\mathbf{v}_j\|_2$.
This baseline exploits concept filtering while keeping the combination weights equal.

\paragraph{Results and discussion.}
We pick 12 pairs of concepts that have the highest co-occurrences, and present the quantitative results in Table~\ref{tab:concept_combination_comparison}. For our method, we first select the optimal $z$ before ranking the image pool with $\mathrm{sim}(\mathbf{x}'_{\mathbf{z}},z\mathbf{v}^1_{j_1}+(1-z)\mathbf{v}^2_{j_2})$.
Although on average, the optimal weighted combination and the average concept combination have similar performances, our methods beats the baseline in more concept pairs.

\begin{table*}[t]
\centering
\caption{mAP@20 for concept combination retrieval on MS COCO using four type of embeddings. \textbf{Weighted}: optimal weighted combination. \textbf{B1}: unfiltered embedding combination. \textbf{B2}: average concept combination. Best scores within each concept pair and embedding model are shown in bold.}
\label{tab:concept_combination_comparison}
\resizebox{\textwidth}{!}{%
\begin{tabular}{cc rrr rrr rrr rrr}
\toprule
\multicolumn{2}{c}{Concept pair}
& \multicolumn{3}{c}{RN50}
& \multicolumn{3}{c}{ViT-B/32}
& \multicolumn{3}{c}{ViT-L/14}
& \multicolumn{3}{c}{DINO} \\
\cmidrule(lr){1-2}
\cmidrule(lr){3-5}
\cmidrule(lr){6-8}
\cmidrule(lr){9-11}
\cmidrule(lr){12-14}
Concept 1 & Concept 2
& Weighted & B1 & B2
& Weighted & B1 & B2
& Weighted & B1 & B2
& Weighted & B1 & B2 \\
\midrule
0 & 1
& 0.459 & 0.277 & \textbf{0.474}
& 0.488 & 0.289 & \textbf{0.503}
& 0.515 & 0.252 & \textbf{0.540}
& 0.516 & 0.269 & \textbf{0.551} \\
0 & 4
& \textbf{0.424} & 0.193 & 0.380
& \textbf{0.460} & 0.189 & 0.408
& \textbf{0.496} & 0.214 & 0.464
& 0.432 & 0.203 & \textbf{0.452} \\
0 & 5
& 0.696 & 0.622 & \textbf{0.760}
& 0.733 & 0.656 & \textbf{0.780}
& 0.787 & 0.673 & \textbf{0.846}
& 0.628 & 0.692 & \textbf{0.753} \\
0 & 6
& \textbf{0.370} & 0.147 & 0.334
& \textbf{0.347} & 0.132 & 0.322
& \textbf{0.417} & 0.107 & 0.412
& \textbf{0.430} & 0.118 & 0.385 \\
0 & 8
& \textbf{0.268} & 0.165 & 0.238
& \textbf{0.301} & 0.132 & 0.260
& \textbf{0.338} & 0.171 & 0.289
& \textbf{0.400} & 0.193 & 0.379 \\
0 & 9
& \textbf{0.426} & 0.155 & 0.406
& 0.407 & 0.167 & \textbf{0.435}
& 0.535 & 0.214 & \textbf{0.591}
& 0.443 & 0.178 & \textbf{0.494} \\
0 & 11
& \textbf{0.214} & 0.121 & 0.188
& 0.183 & 0.113 & \textbf{0.190}
& 0.249 & 0.144 & \textbf{0.259}
& \textbf{0.142} & 0.122 & 0.141 \\
1 & 4
& 0.083 & \textbf{0.105} & 0.072
& 0.120 & 0.092 & \textbf{0.122}
& \textbf{0.142} & 0.078 & 0.109
& 0.129 & 0.089 & \textbf{0.140} \\
6 & 7
& \textbf{0.116} & 0.063 & 0.109
& \textbf{0.141} & 0.076 & 0.124
& \textbf{0.142} & 0.080 & 0.112
& \textbf{0.188} & 0.092 & 0.154 \\
6 & 8
& 0.165 & 0.103 & \textbf{0.182}
& 0.214 & 0.107 & \textbf{0.235}
& \textbf{0.204} & 0.107 & 0.201
& 0.261 & 0.114 & \textbf{0.266} \\
7 & 8
& \textbf{0.160} & 0.119 & 0.139
& \textbf{0.147} & 0.094 & 0.141
& \textbf{0.165} & 0.094 & 0.126
& 0.184 & 0.139 & \textbf{0.207} \\
8 & 11
& 0.133 & 0.070 & \textbf{0.134}
& \textbf{0.128} & 0.081 & 0.127
& \textbf{0.161} & 0.073 & 0.153
& \textbf{0.160} & 0.054 & 0.141 \\
\midrule
\multicolumn{2}{c}{Mean}
& \textbf{0.293} & 0.178 & 0.285
& \textbf{0.306} & 0.177 & 0.304
& \textbf{0.346} & 0.184 & 0.342
& 0.326 & 0.188 & \textbf{0.339} \\
\bottomrule
\end{tabular}%
}
\end{table*}

\end{document}